\documentclass[preprint,12pt]{elsarticle}
\usepackage{amsmath,amsfonts,amssymb}
\usepackage{algorithm}
\usepackage{algpseudocode}
\usepackage{array}
\usepackage{textcomp}
\usepackage{url}
\usepackage{verbatim}
\usepackage{graphicx}
\usepackage{xcolor}
\usepackage{hhline}
\usepackage{booktabs}
\usepackage{tabularx}
\usepackage{adjustbox}
\usepackage{float}
\usepackage{multirow}
\usepackage{makecell}
\usepackage{colortbl}
\definecolor{lightgray}{gray}{0.95}
\usepackage[hidelinks]{hyperref}
\hypersetup{pdftitle={Geometry-Calibrated Closed-Form Shrinkage for Multiplicative Image Denoising},pdfauthor={Xuran Hu, Mingzhe Zhu, Djordje Stankovic, Yujie Zhu, Zhenpeng Feng, Yifang Ban, and Ljubisa Stankovic}}
\biboptions{sort&compress}
\journal{Elsevier}

\begin{document}

\begin{frontmatter}

\title{Closed-Form Nonlocal Shrinkage for Multiplicative Image Denoising and SAR Despeckling}

\author[xidian,kunshan,kth]{Xuran Hu}
\author[xidian,kunshan]{Mingzhe Zhu\corref{cor1}}
\ead{zhumz@mail.xidian.edu.cn}
\author[montenegro]{Djordje Stankovi\'c}
\author[macquarie]{Yujie Zhu}
\author[xidian]{Zhenpeng Feng}
\author[kth]{Yifang Ban}
\author[montenegro]{Ljubi\v{s}a Stankovi\'c}

\cortext[cor1]{Corresponding author.}

\affiliation[xidian]{organization={School of Electronic Engineering, Xidian University},
  city={Xi'an},
  country={China}}
\affiliation[kunshan]{organization={Kunshan Innovation Institute of Xidian University},
  city={Kunshan},
  country={China}}
\affiliation[montenegro]{organization={Department of Electrical Engineering, University of Montenegro},
  city={Podgorica},
  country={Montenegro}}
\affiliation[kth]{organization={Division of Geoinformatics, KTH Royal Institute of Technology},
  city={Stockholm},
  country={Sweden}}
\affiliation[macquarie]{organization={Faculty of Science and Engineering, Macquarie University},
  city={Sydney},
  country={Australia}}

\begin{abstract}
Multiplicative noise poses a challenge in coherent and signal-dependent imaging owing to its intensity-dependent variance and frequently non-Gaussian distribution. We propose a deterministic nonlocal estimator that combines a logarithmic Yeo--Johnson transformation, patch grouping, an adaptive singular basis, and sparse shrinkage. The orthonormal group dictionary makes the weighted Lasso separable and yields an exact coefficient-wise soft-threshold solution. This solution replaces the iterative inner solver and expresses patch reliability and atom importance through a single threshold field. Since the dictionary is estimated from the noisy group, we introduce a random-matrix correction governed by the group aspect ratio $\gamma=p^2/K$. The correction links patch size, group size, and shrinkage strength. Experiments cover gamma-corrupted images from three standard benchmarks and real synthetic aperture radar (SAR) imagery from five sensors. The method gives the best result in 18 of 24 PSNR/SSIM comparisons with twelve published methods and the lowest mean ratio-image deviation across six real SAR configurations. These results support geometry-calibrated nonlocal modeling for structure-preserving image restoration, with SAR despeckling serving as a demanding application. Code is available \href{https://github.com/Teriri1999/Geometry-Calibrated-Closed-Form-Shrinkage-for-SAR-Despeckling}{here}.
\end{abstract}


\end{frontmatter}

\section{Introduction}
Image restoration aims to suppress acquisition noise while preserving edges, textures, and weak structures. This task is particularly challenging under multiplicative noise, whose variance depends on the underlying image intensity and whose distribution is often non-Gaussian. Spatially uniform smoothing may therefore fail to accommodate local noise levels, while excessive smoothing can remove information needed for segmentation, recognition, and quantitative analysis. Synthetic aperture radar (SAR) provides a representative and demanding application. Speckle results from coherent interference and can obscure weak returns and alter the local intensity statistics used for image interpretation \cite{baraha2022systematic,fang2024contrastive,chen2026sds,yang2026glc}. Its statistical properties also depend on the number of looks and the processing chain. These characteristics motivate restoration methods that account for intensity-dependent noise while preserving image structure. Here, we study gamma-distributed multiplicative noise, with SAR despeckling as the principal real-data application.

Nonlocal restoration exploits recurring image structures by grouping similar patches and imposing sparse or low-rank representations. Collaborative filtering, nonlocal sparse coding, and weighted nuclear-norm minimization exemplify this principle \cite{dabov2007bm3d,mairal2009nonlocal,gu2014wnnm,xu2017fncsr,yuan2024lrenss}. These model-based approaches remain useful when paired training data are unavailable, but their performance can depend on regularization weights, patch and group sizes, and iterative solver settings. Learning-based methods instead learn image priors from data, with self-supervised approaches reducing the need for clean targets \cite{lehtinen2018noise2noise,krull2019noise2void,quan2020self2self,wang2022blind2unblind,li2025positive2negative}. Recent SAR networks further exploit attention mechanisms, transformer architectures, and spatial correlations \cite{9633208,perera2022transformer,saha2025cdcfrn}. Their applicability across acquisition conditions may nevertheless be limited by differences in training and test distributions or by the need for image-specific optimization. These concerns motivate a training-free nonlocal estimator with explicit parameter selection and reduced reliance on iterative optimization.

For gamma-distributed multiplicative noise, a logarithmic transformation yields an additive perturbation, while a subsequent Yeo--Johnson mapping is used to reduce distributional asymmetry \cite{ma2024despeckling,yan2025nonlocal}. We combine these transformations with a nonlocal representation in which similar patches are stacked, centered, and projected onto the left singular vectors of the resulting group matrix. This construction has two implications for shrinkage. First, the orthonormal dictionary makes the weighted Lasso separable, yielding an exact coefficient-wise soft-threshold solution that combines patch reliability and atom importance in a single threshold field. Second, because the dictionary is estimated from the noisy group, its basis vectors depend on the noise realization, which must be considered when calibrating the thresholds. We introduce a random-matrix correction governed by the group aspect ratio $\gamma=p^2/K$, where $p$ is the patch width and $K$ is the number of grouped patches. This correction links patch size, group size, and shrinkage strength within the same formulation. Our distributional analysis further indicates that the combined transformation, centering, and singular-vector projection substantially reduce asymmetry at the coefficient shrinkage stage under the tested noise conditions.

We evaluate the proposed training-free estimator on Set12, McMaster, and Kodak24 with simulated gamma-distributed multiplicative noise and on six real SAR configurations from five sensors. The synthetic experiments provide clean references for measuring reconstruction fidelity, while the real SAR experiments assess performance under actual acquisition conditions. In the absence of clean SAR references, we analyze ratio images, formed by dividing each observation by its restored estimate, to assess speckle suppression and the leakage of image structure into the removed component. Together, these experiments evaluate the method across image content, noise levels, and sensors. The main contributions are summarized below.

\begin{enumerate}
\item We show that the orthonormal singular-vector dictionary reduces the weighted Lasso in the proposed nonlocal estimator to exact coefficient-wise soft thresholding. The resulting threshold field unifies patch reliability and atom importance, eliminating the need for an iterative inner solver.
\item We introduce a random-matrix correction to calibrate shrinkage for the data-dependent group dictionary. The correction is governed by the group aspect ratio $\gamma=p^2/K$, linking patch size, group size, and shrinkage strength without per-image calibration.
\item We quantify how transformation, centering, and singular-vector projection change the noise distribution. The empirical analysis shows substantially reduced asymmetry and approximately Gaussian projected noise coefficients under the tested conditions, providing evidence for the noise approximation used at the shrinkage stage.
\item We test the method on three standard image benchmarks with simulated multiplicative noise and on real SAR data from five sensors. These experiments assess reconstruction accuracy and structure preservation using full-reference metrics and ratio-image analysis.
\end{enumerate}

\section{Related Work}

\subsection{Learning-Based Image Restoration}

Supervised image restoration learns a mapping from degraded images to clean references. Residual CNNs provide established baselines for natural-image denoising \cite{zhang2017beyond}, while SAR restoration networks use attention mechanisms, transformer architectures, and correlated dual-channel features to capture spatial dependencies \cite{9633208,perera2022transformer,saha2025cdcfrn}. Since clean SAR references are scarce, supervised training often relies on simulated data, creating a potential mismatch with real acquisitions. Self-supervised methods reduce the need for clean targets by using noisy image pairs or objectives based on blind spots, dropout, and re-visible observations \cite{lehtinen2018noise2noise,krull2019noise2void,quan2020self2self,wang2022blind2unblind,li2025positive2negative}. For SAR despeckling, SAR2SAR uses co-registered multitemporal images \cite{dalsasso2021sar2sar}, whereas Speckle2Void and Speckle2Self learn from single intensity images \cite{molini2021speckle2void,lin2025speckle2self}. Other approaches construct training pairs from real SAR data, combine frequency information with blind-spot learning, or exploit polarimetric and multichannel observations \cite{albisani2025self,chen2026sds,yang2026glc,kato2024polmerlin,denis2025just}. These methods avoid clean training targets, but their requirements vary: some rely on noise independence or suitable masking strategies, while others require additional observations or optimization at test time.

Contrastive learning has been used to reduce the gap between simulated and real SAR data \cite{fang2024contrastive}. Diffusion-based methods have also been explored for conditional generation, posterior sampling, and physics-guided restoration \cite{ma2024despeckling,pan2024sar,hu2024sar,wang2025dps,lu2025dispeckle}, with recent variants aiming to reduce sampling cost or provide control over denoising strength \cite{guo2025efficient,ran2025tunable}. These approaches expand the range of learned restoration models. Our work explores a complementary direction: a training-free nonlocal estimator with a closed-form sparse-coding step and a geometry-based shrinkage correction.

\subsection{Sparse and Nonlocal Image Models}

Sparse denoising represents each image patch using a small number of active dictionary coefficients \cite{aharon2006k,xu2017fncsr}. An $\ell_1$ penalty promotes sparsity, while nonlocal sparse coding jointly represents similar patches to exploit shared image structure \cite{mairal2009nonlocal}. Related approaches operate on groups of similar patches. Collaborative filtering shrinks transform coefficients, whereas weighted nuclear-norm minimization shrinks singular values \cite{dabov2007bm3d,gu2014wnnm}. Joint-prior models further combine low-rank structure with external nonlocal self-similarity \cite{yuan2024lrenss}, and theoretical recovery guarantees have been established for certain low-rank denoisers within plug-and-play formulations \cite{gavaskar2023pnp}. For multiplicative noise, nonlocal rank minimization applies group-based low-rank modeling in the logarithmic domain \cite{yan2025nonlocal}.

For coherent imaging, restoration models should account for the statistical properties of multiplicative noise. Logarithmic and variance-stabilizing transformations allow filtering in a transformed domain, and nonlocal and multichannel methods adapt Gaussian denoising techniques to speckle \cite{parrilli2012sar,deledalle2017mulog}. Related approaches use nonlocal matrix decomposition in the logarithmic domain to preserve sparse structures in SAR time series \cite{kang2023logsar}, or explicitly seek to preserve the theoretical speckle distribution \cite{li2020statistical}. Low-rank despeckling methods also exploit the shared structure of grouped patches \cite{xu2023edge,yan2025nonlocal}. Dictionary-based and plug-and-play approaches include methods that use iterative shrinkage \cite{10032489,baraha2022systematic}. In particular, weighted sparse recovery combines homomorphic transformation, nonlocal grouping, adaptive dictionaries, and coefficient-dependent regularization \cite{9484779}. Our method builds on this framework by using an orthonormal singular-vector dictionary to obtain a closed-form sparse-coding solution and relating the shrinkage correction to patch-group geometry.

Dictionary-based methods often transform multiplicative observations to simplify noise modeling and sparse recovery \cite{liu2017over}. For a centered nonlocal group with singular value decomposition $Y=U\Sigma V^\top$, choosing the orthonormal dictionary $D=U$ makes the quadratic fidelity term separable in the coefficient domain. However, because $U$ is estimated from the noisy group, it cannot be treated as a fixed basis independent of the noise. This dependence motivates a correction to the noise scale used for thresholding. We use the random-matrix spectral edge to motivate a correction governed by the group aspect ratio, providing a geometry-based rule for setting the shrinkage strength.

\section{Proposed Method}

\subsection{Overview of Proposed Model}

\begin{figure}[t]
	\centering
	\includegraphics[width=\textwidth]{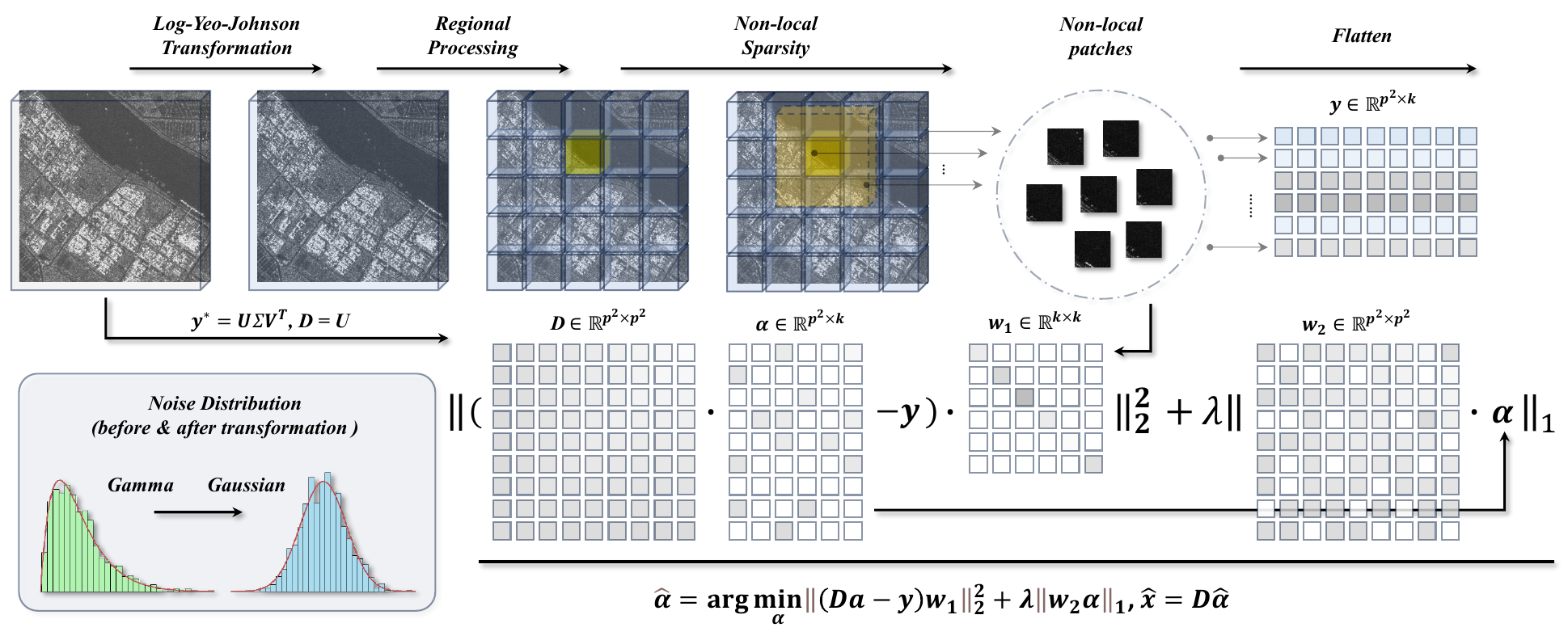}
	\caption{Overview of the proposed image-restoration pipeline. The logarithmic Yeo--Johnson transformation reduces the asymmetry of multiplicative noise. Similar patches are grouped and represented in their orthonormal singular basis $D=U$, which yields an analytic coefficient-wise threshold.}
	\label{fig_1}
\end{figure}

The model combines a statistical transformation with region-aware nonlocal sparse recovery, as shown in Fig. \ref{fig_1}. For gamma-distributed multiplicative noise, a logarithmic Yeo--Johnson transformation produces an approximately additive Gaussian field. Patches are grouped by Euclidean similarity and coded using factors that describe local reliability and the adaptive sparsity prior. The orthonormal group dictionary permits an exact solution of the resulting weighted Lasso. The noise shape and group geometry determine parameters that would require tuning.

\subsection{Statistical Transformation of Multiplicative Noise}

We model the observed intensity as $y=xn$, where $x$ is the noise-free image and $n\sim\Gamma(L,L)$ is unit-mean multiplicative noise. In coherent imaging, $L$ is the equivalent number of looks; more generally, it is the gamma shape parameter that controls the noise variance.

\begin{equation}
	p(n)=\frac{L^L n^{L-1} e^{-L n}}{\Gamma(L)}
\end{equation}
The logarithm converts the multiplicative noise into an additive term. A Yeo--Johnson mapping \cite{ma2024despeckling} then reduces its skewness.

\begin{equation}
	\operatorname{Yeo}_{\lambda}(z)=
	\begin{cases}
	\big((z+1)^\lambda-1\big)/\lambda, & z\geq0,\ \lambda\neq0,\\
	\log(z+1), & z\geq0,\ \lambda=0,\\
	-\big((1-z)^{2-\lambda}-1\big)/(2-\lambda), & z<0,\ \lambda\neq2,\\
	-\log(1-z), & z<0,\ \lambda=2.
	\end{cases}
\end{equation}
The parameter $\lambda=\lambda^*(L)$ is determined offline from pure $\Gamma(L,L)$ noise by minimizing the sum of absolute skewness and absolute excess kurtosis over $\{0,0.01,\ldots,4\}$ using $4\times10^5$ samples. The fixed values are listed within the implementation details.

The analytic inverse applied before exponentiation is

\begin{equation}
	\operatorname{Yeo}_{\lambda}^{-1}(q)=
	\begin{cases}
	(\lambda q+1)^{1/\lambda}-1, & q\geq0,\ \lambda\neq0,\\
	e^q-1, & q\geq0,\ \lambda=0,\\
	1-\big(1-(2-\lambda)q\big)^{1/(2-\lambda)}, & q<0,\ \lambda\neq2,\\
	1-e^{-q}, & q<0,\ \lambda=2.
	\end{cases}
	\label{eq:yeo-inverse}
\end{equation}

The transform changes only the domain used for grouping and shrinkage. After applying \eqref{eq:yeo-inverse} and exponentiating, we multiply the estimate by the ratio of the observation mean to the estimated mean. This empirical mean-preserving step addresses log-domain radiometric bias without requiring a clean reference.

\subsection{Nonlocal Sparse Image Reconstruction}

Sparse reconstruction represents a patch using a few dictionary coefficients. We extend this model to $K$ similar patches, selected by squared Euclidean distance and stacked in $Y\in\mathbb R^{p^2\times K}$. The centered group's left singular vectors define $D$; replacing the sparsity count with an $\ell_1$ penalty yields the weighted Lasso derived below.

\paragraph{Grouping}
At high noise levels, patch distances measured directly from the observation are dominated by multiplicative fluctuations, and the nearest observed patch may have a different latent structure. We therefore match patches using a Gaussian-smoothed version of the current estimate but stack and reconstruct the corresponding unsmoothed patches. Smoothing affects only group membership; no smoothed value enters $Y$, its singular basis, or aggregation. Over bandwidths $b\in[0,1]$, synthetic PSNR changes by at most $0.4$~dB. We fix $b=1$ for synthetic data and $b=0$ for real data without selecting it from the observed image.

\begin{algorithm}[t]
	\caption{Geometry-calibrated multiplicative image denoising.}
	\label{alg:image_denoising}
	\begin{algorithmic}[1]
		\State \textbf{Input:} $y,L,p,K,b,\eta,r,T_{\max}$; \textbf{output:} $\hat{x}$
		\State $\gamma\gets p^2/K$, $\lambda\gets\lambda^*(L)$, $c\gets1.5$ if $L=1$; otherwise $c\gets c^*(\gamma,L)$, $y_{\rm tr}\gets\operatorname{Yeo}(\log y;\lambda)$
		\State Estimate $\hat\sigma_0$ from the patch-covariance spectrum; set $\sigma_{\min}\gets\eta\hat\sigma_0$, $\hat{x}^{(0)}\gets y_{\rm tr}$, and $t\gets0$
		\While{$t < T_{\max}$}
		\State $\tilde\sigma_k\gets\big|\hat\sigma_0^2-\mathrm{mean}_i(y_{{\rm tr},ik}-\hat{x}^{(t)}_{ik})^2\big|^{1/2}$, $\hat\sigma_k\gets\max(c\tilde\sigma_k,\sigma_{\min})$
		\State Smooth $\hat{x}^{(t)}$ with bandwidth $b$ to obtain $\hat{x}_b^{(t)}$
		\For{each reference patch}
		\State Refresh matches every two iterations on $\hat{x}_b^{(t)}$, stack the unsmoothed patches as $Y$, and set $\bar y\gets Y\mathbf1/K$, $Y_c\gets Y-\bar y\mathbf1^\top$
		\State Obtain $D,\Sigma$ from the $K\times K$ Gram matrix and set $B\gets D^\top Y_c$
		\State $\hat{\alpha}_{ik} \gets \mathcal{S}\big( B_{ik},\; \hat{\sigma}_k^{2} / S_i \big)$
		\State Apply \eqref{eq:rank}; reconstruct $\hat Y\gets D\hat\alpha+\bar y\mathbf1^\top$
		\EndFor
		\State Aggregate $\hat Y$ with weights $1/\hat\sigma_k$ into $\hat{x}^{(t+1)}$; set $t\gets t+1$
		\EndWhile
		\State $\hat{x} \gets \exp(\operatorname{Yeo}^{-1}(\hat{x}^{(T_{\max})}))$; rescale to the mean of $y$
	\end{algorithmic}
\end{algorithm}

\subsection{Sparsity-guided Posterior Estimation}

We use a Gaussian likelihood and a Laplace prior, holding the group-derived dictionary and prior scales fixed during sparse coding.

For an uncentered group $Y\in\mathbb{R}^{d\times K}$, $d=p^2$, define
\begin{equation}
 \bar y=K^{-1}Y\mathbf{1},\qquad Y_c=Y-\bar y\mathbf{1}^\top
 =U_{\varrho}\Sigma_{\varrho}V_{\varrho}^\top,\qquad D=U_{\varrho},
 \label{eq:centered-group}
\end{equation}
where $\varrho=\operatorname{rank}(Y_c)\le\min(d,K-1)$, $\Sigma_{\varrho}=\operatorname{diag}(S_1,\ldots,S_{\varrho})$ contains positive singular values, and $D^\top D=I_{\varrho}$. Below, $y_k$ denotes column $k$ of $Y_c$, and $\alpha\in\mathbb{R}^{\varrho\times K}$. Zero singular directions are excluded; if $\varrho=0$, reconstruction returns $\bar y\mathbf{1}^\top$. Numerically, directions with $S_i\le10^{-7}S_1$ are omitted, and the implementation uses $1/(S_i+10^{-12})$ to stabilize reciprocal singular values. The formulas below use $1/S_i$ for the ideal positive-spectrum model; the guarded weights retain the same closed-form solution with the corresponding adjusted threshold.

The restored mean has noise variance $K^{-2}\sum_{k,\ell}\operatorname{Cov}(\varepsilon_k,\varepsilon_\ell)$ at each patch position, reducing to $\sigma^2/K$ for independent, equal-variance noise.

\paragraph{Likelihood}
We approximate the centered perturbation as Gaussian with constant variance within each patch. The product likelihood neglects correlations induced by centering, overlap, and data-dependent grouping.

\begin{equation}
	P(Y_c \mid \alpha)=\prod_{k=1}^K\left(2 \pi \sigma_k^2\right)^{-p^2/2} e^{-\frac{1}{2 \sigma_k^2}\left\|y_k-D \alpha_k\right\|_2^2}
\end{equation}
$\alpha_k$ is the $k$th coefficient column, and $\sigma_k$ is the noise level of patch $y_k$. This formulation allows the noise level to vary across patches.

\paragraph{Prior}
A sparse $\alpha$ contains few nonzero entries. We therefore use an independent Laplace prior for its coefficients.

\begin{equation}
	P\left(\alpha_{i k}\right)=\frac{1}{2 S_i} e^{\left(-\frac{\left|\alpha_{i k}\right|}{S_i}\right)}
	\label{eq:laplace}
\end{equation}
The scale $S_i$ is the singular value associated with the $i$th atom. Atoms that contain more of the group energy consequently receive a smaller penalty and can have larger coefficients.

\paragraph{Objective function}
Under the independence assumption, $P(\alpha)$ is the product of \eqref{eq:laplace} over $i$ and $k$. Taking the negative logarithm of the posterior and discarding terms that do not depend on $\alpha$ gives

\begin{equation}
	\hat{\alpha} 
	= \arg\min_{\alpha}\, 
	\sum_{k=1}^{K} \frac{1}{2\sigma_k^2}\,\lVert y_k - D\alpha_k \rVert_2^2 
	+ \sum_{k=1}^{K}\sum_{i=1}^{\varrho} S_i^{-1}\,|\alpha_{ik}|
	\label{eq:map-simplified}
\end{equation}

Let $w_1 = \mathrm{diag}(\hat{\sigma}_1^{-1}, \dots, \hat{\sigma}_K^{-1})$, where the per-patch estimate $\hat{\sigma}_k$ in \eqref{eq:sigmahat} replaces the unknown $\sigma_k$, and let $w_2 = \Sigma_{\varrho}^{-1}$. Equation \eqref{eq:map-simplified} can then be written as

\begin{equation}
	\hat{\alpha}=\arg \min _\alpha \tfrac{1}{2}\left\|(D \alpha-Y_c) w_1\right\|_F^2+\left\|w_2 \alpha\right\|_1
\end{equation}

\paragraph{Closed-form solution of the weighted Lasso}
The data-adaptive dictionary $D$ consists of left singular vectors, with $D^\top D=I_{\varrho}$. This property gives an exact solution to the weighted Lasso and removes the need for an iterative solver. The identity $\left\|y_k-D \alpha_k\right\|_2^2=\left\|D^\top y_k-\alpha_k\right\|_2^2+\left\|\left(I-D D^\top\right) y_k\right\|_2^2$, together with the independence of the second term from $\alpha$, makes the objective in \eqref{eq:map-simplified} separable over the entries of $\alpha$. With $B=D^\top Y_c$, each entry solves the scalar problem:

\begin{equation}
	\hat{\alpha}_{i k}=\arg \min _a \frac{1}{2 \hat{\sigma}_k^2}\left(a-B_{i k}\right)^2+S_i^{-1}\left|a\right|
\end{equation}
whose minimizer is the soft-thresholding operator

\begin{equation}
	\hat{\alpha}_{i k}=\mathcal{S}\left(B_{i k}, \tau_{i k}\right), \quad \tau_{i k}=\frac{\hat{\sigma}_k^2}{S_i}=\frac{w_{2, i}}{w_{1, k}^2}
	\label{eq:closed-form}
\end{equation}

Here $\mathcal{S}(u,\tau)=\operatorname{sign}(u)\max(|u|-\tau,0)$ for $\tau\ge0$. The exact solution is conditional on the fixed dictionary and positive weights, including when $D$ is rectangular.

The update removes the iterative inner solver and its convergence parameters. Both weights enter through $\tau_{ik}$: $w_1$ describes patch reliability and $w_2$ atom importance. Table \ref{tb13} isolates their adaptivity by fixing each factor at its group mean.

\paragraph{Rank selection}
A leading singular direction in a flat group can be noise-dominated yet receive little shrinkage because $\tau_{ik}\propto S_i^{-1}$. We therefore retain

\begin{equation}
	\mathcal I=\big\{i:S_i>r\,\max(\tilde\sigma_{\rm ref},\sigma_{\min})\sqrt K\big\}
	\label{eq:rank}
\end{equation}
and set coefficients outside $\mathcal I$ to zero. Here $\tilde\sigma_{\rm ref}$ is the reference patch's uncorrected noise estimate. Masked soft thresholding exactly solves the weighted objective with these additional zero constraints. The empirical parameter $r$ balances noise rejection and weak-structure retention; the cutoff is not a theoretical spectral edge. Using the uncorrected scale separates screening from the geometry correction $c$.

\paragraph{Reconstruction and aggregation}
Each group is reconstructed as $\hat Y=D\hat\alpha+\bar y\mathbf1^\top$. Let $R_{gk}$ extract patch $k$ of group $g$ from the image and let $\omega_{gk}=1/\hat\sigma_{gk}$. Overlapping estimates are combined by
\begin{equation}
 \hat x^{(t+1)}=\frac{\sum_{g,k}\omega_{gk}R_{gk}^\top\hat y_{gk}}
 {\sum_{g,k}\omega_{gk}R_{gk}^\top\mathbf1},
 \label{eq:aggregation}
\end{equation}
with elementwise division; uncovered pixels retain their previous estimate. For $K\le d$, forming and diagonalizing a group Gram matrix costs $O(dK^2+K^3)$, while thresholding costs $O(\varrho K)$ once the projected coefficients are available. These are per-group costs; matching, reconstruction, aggregation, and repeated outer iterations also contribute to total runtime.

The resulting procedure is outlined in Algorithm \ref{alg:image_denoising}.

\subsection{Adaptive Noise Estimation}
\label{sec:sigma}
For $n\sim\Gamma(L,L)$, let $\mu_L=\psi(L)-\log L$ and $\delta=\log n-\mu_L$, so that $\mathbb E\delta=0$ and $\operatorname{Var}(\delta)=\psi'(L)$. With $g=\operatorname{Yeo}_{\lambda}$ and $z_0=\log x+\mu_L$, a local expansion gives
\begin{equation}
 \operatorname{Var}[g(\log y)]\approx[g'(z_0)]^2\psi'(L),\qquad
 g'(z)=\begin{cases}(1+z)^{\lambda-1},&z\ge0,\\(1-z)^{1-\lambda},&z<0.\end{cases}
 \label{eq:local-variance}
\end{equation}
Thus, unlike logarithmic noise, the transformed perturbation is generally signal-dependent. This first-order approximation motivates patch-dependent noise scales; it does not establish Gaussianity or exact bias removal after inversion. We use a global initialization, patch-level refinement, and a geometry-dependent correction.

\paragraph{Global initialization}
We estimate $\hat\sigma_0$ from the ordered eigenvalues $\nu_1\le\cdots\le\nu_{p^2}$ of the transformed-image patch covariance, sampled on a subgrid. We average the largest lower-spectrum subset with equal counts above and below its mean:

\begin{equation}
\begin{aligned}
	\hat{\sigma}_0^2 &= \frac{1}{n}\sum_{j=1}^{n} \nu_j,\\
	n &= \max \left\{m:\left|\{j\le m:\nu_j>\bar{\nu}_m\}\right|\right.\\
	& \left.=\left|\{j\le m:\nu_j<\bar{\nu}_m\}\right|\right\}
\end{aligned}
	\label{eq:sigma0}
\end{equation}
Here $\bar\nu_m$ is the mean of the smallest $m$ eigenvalues. The estimate assumes a noise-dominated lower spectrum and can fail when structure occupies all scales. The floor $\sigma_{\min}=\eta\hat\sigma_0$, $\eta=0.2$, prevents nearly zero patch estimates from dominating aggregation through their reciprocal weights.

\begin{figure}[t]
	\centering
	\includegraphics[width=\linewidth]{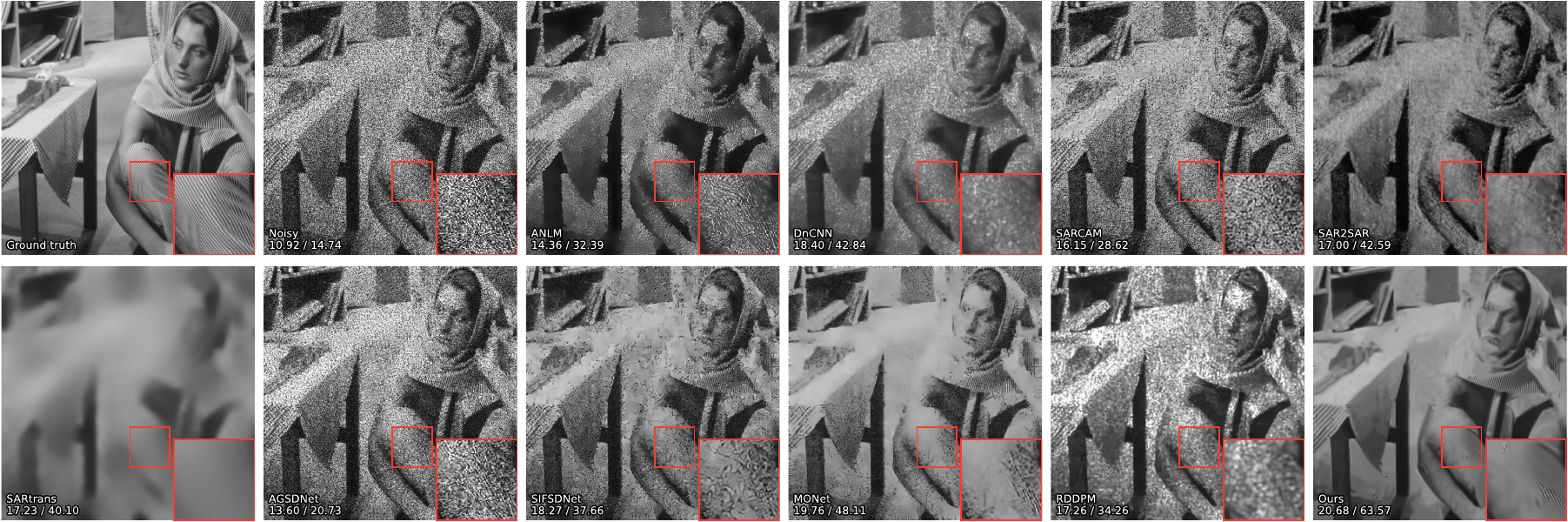}
	\caption{One-look Set12 results. Panels are ordered as labeled. Insets show the marked regions, with PSNR/SSIM below. DIP, CL-SAR, and S3DIP used different half-resolution scenes and are omitted.}
	\label{fig_2}
\end{figure}

\begin{figure}[t]
	\centering
	\includegraphics[width=\linewidth]{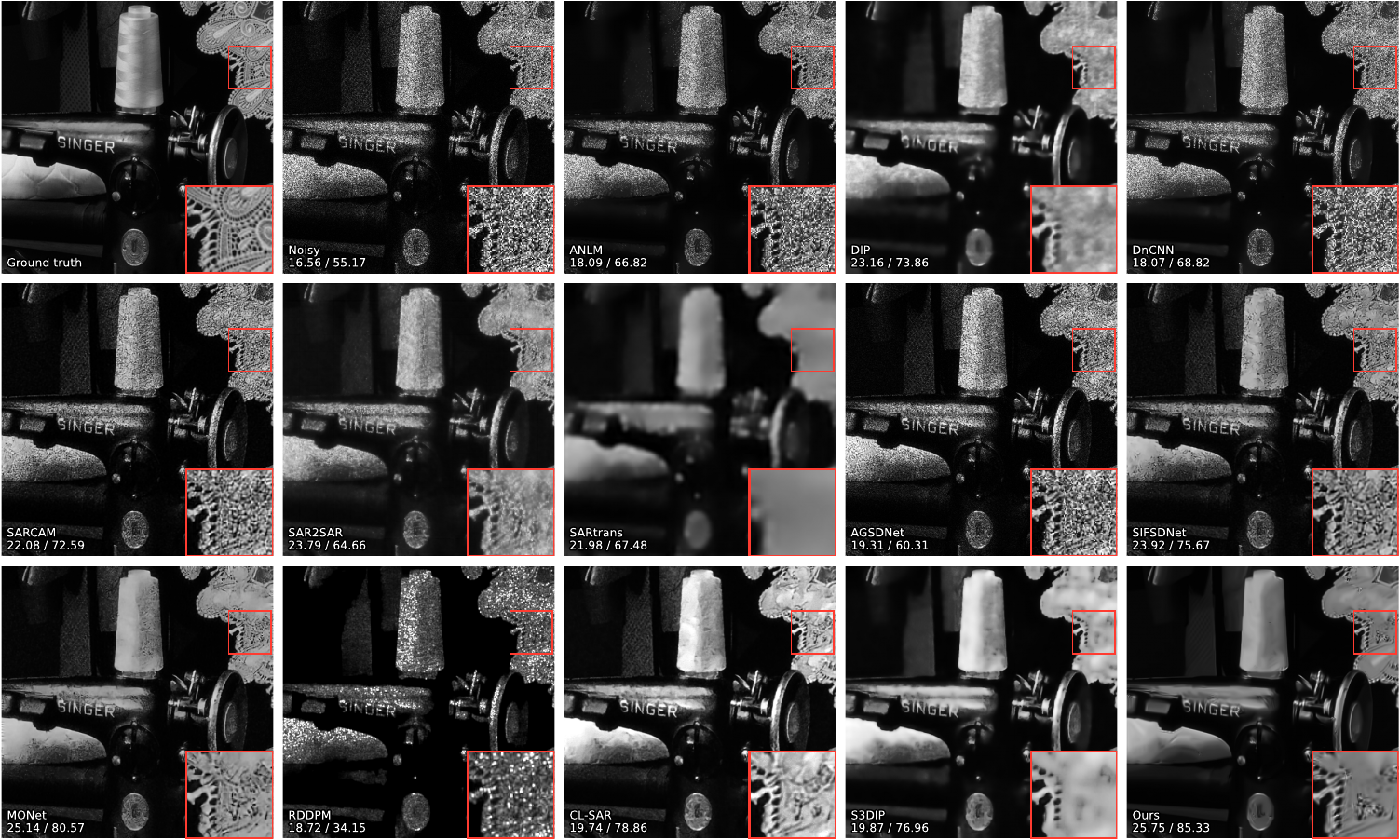}
	\caption{Two-look McMaster results. Panels are ordered as labeled. Insets show the marked regions, with PSNR/SSIM below.}
	\label{fig_3}
\end{figure}

\begin{figure}[t]
	\centering
	\includegraphics[width=\linewidth]{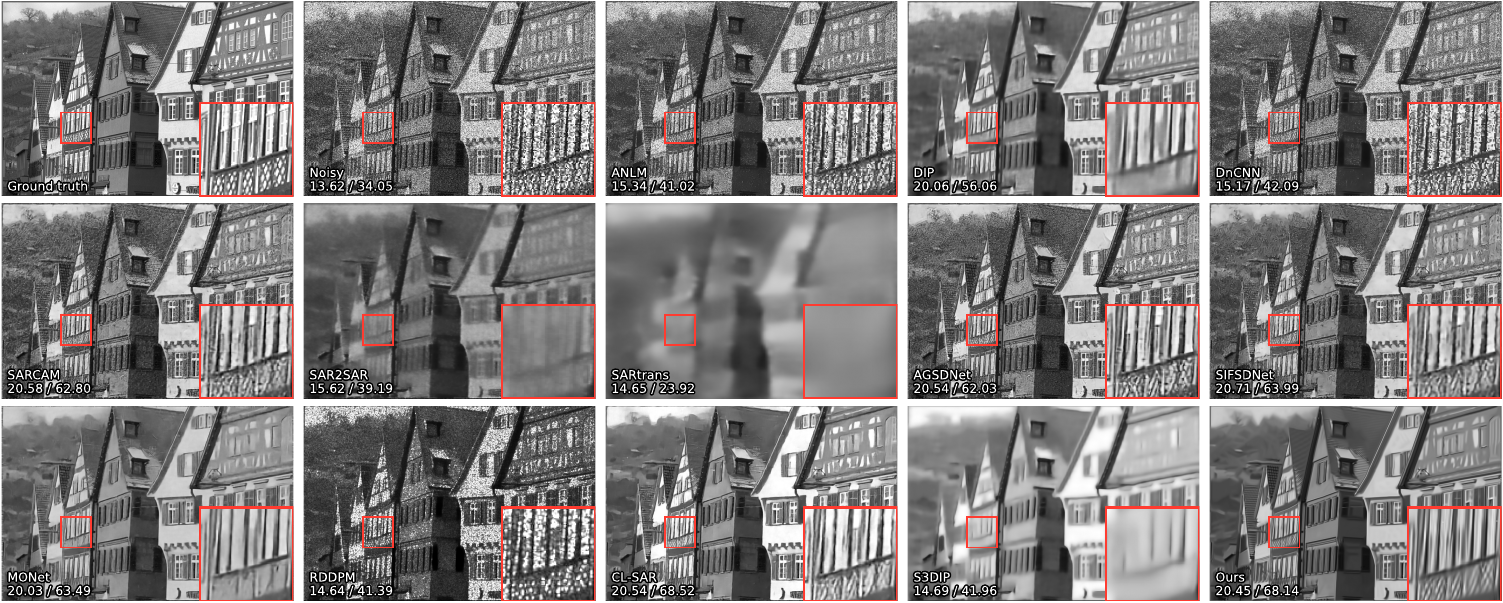}
	\caption{Four-look Kodak24 comparison. Panels are ordered as labeled, and insets show the marked regions. PSNR in decibels and SSIM multiplied by 100 are shown below each panel.}
	\label{fig_kodak}
\end{figure}

\paragraph{Per-patch refinement by variance differencing}
For the transformed observation $\tilde y$ and current estimate $\hat x^{(t)}$, we estimate remaining noise by residual-power differencing:

\begin{equation}
	\tilde{\sigma}_k^{(t)} = \Big| \hat{\sigma}_0^2 - \mathrm{mean}_i \big( \tilde{y}_{ik} - \hat{x}^{(t)}_{ik} \big)^2 \Big|^{1/2}, \quad k=1,\dots,K,
	\label{eq:sigmak}
\end{equation}
For an additive working model $\tilde y=x_{\rm tr}+\varepsilon$ and estimation error $e=\hat x^{(t)}-x_{\rm tr}$, the identity $\mathbb E(\tilde y-\hat x^{(t)})^2=\mathbb E\varepsilon^2+\mathbb E e^2-2\mathbb E(\varepsilon e)$ shows that residual power also depends on reconstruction error and its correlation with noise. Equation \eqref{eq:sigmak} is therefore not an unbiased variance identity. The absolute value is an implementation safeguard when the measured residual exceeds the global power; the floor in \eqref{eq:sigmahat} prevents nearly zero estimates from dominating aggregation.

\paragraph{Geometry-dependent correction}
The per-patch noise level used in the threshold is

\begin{equation}
	\hat{\sigma}_k = \max(c\,\tilde{\sigma}_k,\ \sigma_{\min}),
	\label{eq:sigmahat}
\end{equation}
The scale factor $c$ changes the threshold quadratically and accounts for noise concentration in data-dependent singular directions. For independent white noise, the asymptotic spectral edge is $\sigma\sqrt K(1+\sqrt\gamma)$, $\gamma=p^2/K$ \cite{gavish2014optimal}. A pure-noise fit in $1+\sqrt\gamma$ gives $R^2=0.996$ over $\gamma\in[0.4,14.4]$.

To relate this edge to coefficient energy, consider $N_c=N(I-\mathbf1\mathbf1^\top/K)=U_N\Sigma_NV_N^\top$ and $B_N=U_N^\top N_c$. Exactly,
\begin{equation}
 K^{-1}\sum_{k=1}^K(B_N)_{ik}^2=S_i(N_c)^2/K.
 \label{eq:noise-row-energy}
\end{equation}
For independent Gaussian entries of $N$ with variance $\sigma^2$, centering leaves $K-1$ independent column directions, giving the asymptotic leading-row RMS scale $\sigma(\sqrt d+\sqrt{K-1})/\sqrt K$. Its large-$K$ form is $\sigma(1+\sqrt\gamma)$. This pure-noise argument motivates the functional dependence, but does not derive optimal thresholds for a signal-containing group or the fitted coefficients below. Overlap and adaptive matching introduce further departures from the independent-noise model.

We therefore calibrate $c$ using the same functional form.

\begin{equation}
	c^{*}(\gamma, L) = a(L)\,\big(1+\sqrt{\gamma}\,\big) + b(L), \qquad \gamma = p^2/K
	\label{eq:cstar}
\end{equation}
The released implementation uses $a(L)=0.3497+0.0580\log_2L$ and $b(L)=0.5580-0.2520\log_2L$, obtained by linear regression in $\log_2L$ using fits at $L=2,4,8$. Across twelve $(p,K)$ combinations with $\gamma$ from $0.8$ to $25.6$, \eqref{eq:cstar} yields $R^2=0.90$, $0.97$, and $0.98$ for $L=2$, $4$, and $8$, respectively. Fits obtained with $\gamma\le14.4$ predict the optimum at $\gamma=25.6$ within $0.16$. The correction therefore couples patch size and group size through their aspect ratio. A constant calibrated for one geometry should not be transferred to another.

\paragraph{One-look case}
Equation \eqref{eq:cstar} is calibrated at $L=2,4,8$. The implementation evaluates the same expression for $L>8$ by extrapolation, without clipping; these values lie outside the fitted look range. At one look, the optimum no longer follows the aspect ratio and moves toward stronger shrinkage without a clear turning point. We therefore use the fixed experimental value $c=1.5$ instead of extrapolating \eqref{eq:cstar}.

Pure-noise theory motivates the geometry dependence; image experiments calibrate its coefficients. No per-image refitting is performed. Table \ref{tb14} tests transfer across patch-group geometries. The correction changes the fidelity scale, leaving singular values unchanged.

\section{Experiments}

\subsection{Experimental Setup}

\paragraph{Datasets}
We use the grayscale Set12, McMaster, and Kodak24 benchmarks to evaluate multiplicative image denoising under controlled conditions. Each clean image is multiplied by an independent $\Gamma(L,L)$ realization for $L\in\{1,2,4,8\}$. The clean image is used only for evaluation. To test the same estimator outside simulation, we additionally use real SAR data from five sensors.

The real data comprise Sentinel-1 \cite{torres2012gmes}, Gaofen-3 \cite{zhao2021china}, TerraSAR-X, miniSAR, and FARAD X- and Ka-band images. The six configurations range in spatial resolution from $5$--$25$~m to $0.1$~m, as summarized in Table \ref{tb1}.

\begin{table}[!t]\setlength{\belowcaptionskip}{6pt}
	\scriptsize
	\centering
	\caption{Characteristics of the real SAR datasets.\label{tb1}}
	\begin{adjustbox}{max width=0.82\linewidth}
	\begin{tabular}{lccc}
		\toprule
		\textbf{Sensor} & \textbf{Band} & \textbf{Size} & \textbf{Resolution (m)} \\
		\midrule
		Sentinel-1 & C & 256$\times$256 & 5--25 \\
		Gaofen-3   & C & 512$\times$512 & 3--25 \\
		TerraSAR-X & X & 512$\times$512 & 3 \\
		miniSAR & X & 1255$\times$819 & 0.1 \\
		\multirow{2}{*}{FARAD} & X & 1434$\times$1007 & 0.1 \\
		& Ka & 650$\times$1020 & 0.1 \\
		\bottomrule
	\end{tabular}
	\end{adjustbox}
\end{table}

\paragraph{Evaluation metrics}
Synthetic experiments use PSNR in decibels and SSIM. Bold and underlined entries indicate the best and second-best results. Real SAR images have no clean reference, so we evaluate $r=y/\hat{x}$ using $\Delta=|\bar r-1|+|L s_r^2-1|+\bar\rho_r$, where $\bar\rho_r$ is the mean absolute normalized autocorrelation over the eight nonzero lags in a $3\times3$ neighborhood. All terms approach zero when the ratio contains only the modeled multiplicative noise.

\paragraph{Implementation details}
No parameter is tuned for an individual image or dataset. We use $c=1.5$ at one look and $c=c^{*}(\gamma,L)$ otherwise, with extrapolation beyond $L=8$. For $L=1,2,3,4,6,8,12,16$, the fixed $\lambda^*(L)$ values are $1.51,1.48,1.47,1.46,1.44,1.43,1.42,1.42$. For real data, local ENL is computed in $32\times32$ windows with stride 8, and its 95th percentile is mapped logarithmically to the nearest candidate look number. We use $p=10$, $K=20$, $b=1$ for synthetic data, $b=0$ for real data, $r=1.5$ except $r=1$ at one look, $\eta=0.2$, and $T_{\max}=12$. Reference patches have stride 3; matching uses a search half-window of 20 pixels and is refreshed every two outer iterations. The estimator is deterministic.

\paragraph{Baselines}
We compare twelve methods using the official settings and pretrained models released by their authors. The methods are ANLM \cite{xiao2020asymptotic}, DIP \cite{ulyanov2018deep}, DnCNN \cite{zhang2017beyond}, SAR2SAR \cite{dalsasso2021sar2sar}, SARCAM \cite{9633208}, SARtrans \cite{perera2022transformer}, AGSDNet \cite{thakur2022agsdnet}, SIFSDNet \cite{thakur2022sifsdnet}, MONet \cite{vitale2023sar}, CL-SAR \cite{fang2024contrastive}, S3DIP \cite{albisani2025self}, and RDDPM \cite{hu2024sar}.

\begin{figure}[t]
	\centering
	\includegraphics[width=\textwidth]{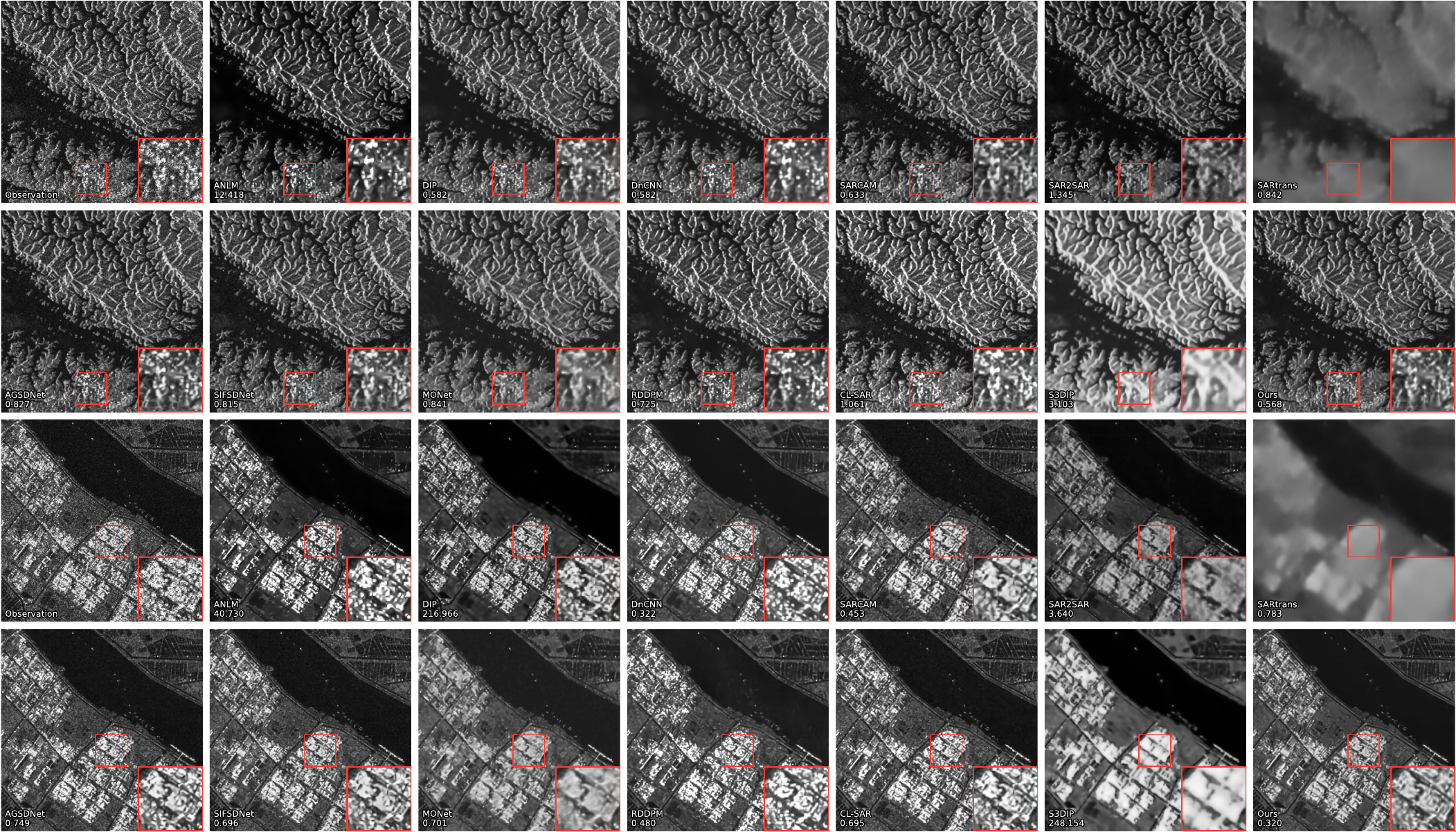}
	\caption{Sentinel-1 results on two scenes. Panels are ordered as labeled, and insets show the marked regions. Values below the panels are the ratio-image deviations from Table \ref{tb6}. Lower values are better.}
	\label{fig_6}
\end{figure}

\newcommand{\realresultstable}{%
\begin{table}[!t]\setlength{\belowcaptionskip}{6pt}
	\scriptsize
	\centering
	\caption{Ratio-image evaluation on real SAR data. The total deviation measures departures from $\mathrm{E}[G]=1$, $L\mathrm{Var}[r]=1$, and zero autocorrelation. The value of $L$ is estimated for each image. Lower values are better.\label{tb6}}
	\setlength{\tabcolsep}{4pt}
	\begin{adjustbox}{max width=\linewidth}
	\begin{tabular}{>{\raggedright\arraybackslash}p{1.9cm}*{10}{c}}
		\toprule[1pt]
		\multirow{2}{*}{\textbf{Method}} & \multicolumn{6}{c}{\textbf{Total deviation per configuration}} & \multicolumn{3}{c}{\textbf{Mean of each term}} & \multirow{2}{*}{\textbf{Mean}} \\
		\cmidrule(lr){2-7} \cmidrule(lr){8-10}
		& FARAD-Ka & FARAD-X & Gaofen-3 & Sentinel-1 & TerraSAR-X & miniSAR & $|\mathrm{E}[G]{-}1|$ & $|L\mathrm{Var}[r]{-}1|$ & autocorr. & \\
		\midrule
		ANLM\cite{xiao2020asymptotic} & 0.763 & 0.921 & 0.941 & 19.528 & 0.519 & 12.996 & 0.249 & 5.297 & 0.399 & 5.945 \\
		DIP\cite{ulyanov2018deep} & 1.152 & 4.631 & 6.246 & 56.450 & 23.649 & 5.114 & 0.569 & 15.079 & 0.560 & 16.207 \\
		DnCNN\cite{zhang2017beyond} & 0.613 & 0.708 & 0.602 & \underline{0.439} & 0.618 & 0.784 & 0.047 & \underline{0.320} & 0.261 & 0.627 \\
		SARCAM\cite{9633208} & 0.673 & 1.618 & \underline{0.503} & 0.525 & \underline{0.517} & 1.143 & 0.057 & 0.625 & 0.148 & 0.830 \\
		SAR2SAR\cite{dalsasso2021sar2sar} & 1.242 & 1.272 & 2.429 & 2.622 & 1.550 & 3.402 & 0.379 & 1.260 & 0.448 & 2.086 \\
		SARtrans\cite{perera2022transformer} & 2.171 & 1.862 & 0.734 & 0.780 & 0.792 & 1.947 & 0.126 & 0.720 & 0.536 & 1.381 \\
		AGSDNet\cite{thakur2022agsdnet} & 0.859 & 0.868 & 0.955 & 0.780 & 0.915 & 0.763 & \underline{0.020} & 0.736 & \textbf{0.100} & 0.856 \\
		SIFSDNet\cite{thakur2022sifsdnet} & 0.557 & \textbf{0.488} & 0.786 & 0.737 & 0.683 & \underline{0.370} & 0.080 & 0.385 & 0.139 & \underline{0.603} \\
		MONet\cite{vitale2023sar} & \underline{0.515} & 0.545 & 0.767 & 0.719 & 0.737 & \textbf{0.356} & 0.117 & 0.333 & 0.156 & 0.607 \\
		RDDPM\cite{hu2024sar} & 216.721 & 192.839 & 38.576 & 0.523 & 90.075 & 310.118 & 2.272 & 138.630 & 0.574 & 141.475 \\
		CL-SAR\cite{fang2024contrastive} & 0.918 & 0.940 & 1.003 & 0.826 & 1.006 & 0.505 & 0.094 & 0.639 & \underline{0.134} & 0.867 \\
		S3DIP\cite{albisani2025self} & 1.945 & 2.042 & 5.164 & 56.857 & 3.723 & 10.715 & 0.507 & 12.321 & 0.579 & 13.407 \\
		Ours & \textbf{0.321} & \underline{0.491} & \textbf{0.422} & \textbf{0.422} & \textbf{0.509} & 0.432 & \textbf{0.020} & \textbf{0.210} & 0.203 & \textbf{0.433} \\
		\bottomrule[1pt]
	\end{tabular}
	\end{adjustbox}
\end{table}
}

\subsection{Results on Synthetic Multiplicative-Noise}

McMaster and Kodak24 are converted to grayscale and corrupted at four gamma noise levels. Figures \ref{fig_2} and \ref{fig_3} show representative one- and two-look results for Set12 and McMaster, and Fig. \ref{fig_kodak} gives the four-look Kodak24 comparison. Although the noise model is motivated by coherent imaging, these benchmarks contain generic edges, textures, and repeated patterns. They therefore isolate the behavior of the image prior from sensor-specific scene content. The proposed method suppresses multiplicative fluctuations while retaining fine structures across all three datasets.

\begin{table}[!t]\setlength{\belowcaptionskip}{6pt}
	\tiny
	\centering
	\caption{Synthetic image-denoising results under gamma-distributed multiplicative noise (PSNR/SSIM).\label{tb2}}
	\setlength{\tabcolsep}{0.8pt}
	\renewcommand{\arraystretch}{0.82}
	\begin{tabularx}{0.8\linewidth}{@{}p{1.38cm}*{8}{>{\centering\arraybackslash}X}@{}}
		\toprule[1pt]
		\multirow{2}{*}{\textbf{Method}} & \multicolumn{2}{c}{\textbf{1-Look}} & \multicolumn{2}{c}{\textbf{2-Look}} & \multicolumn{2}{c}{\textbf{4-Look}} & \multicolumn{2}{c}{\textbf{8-Look}} \\
		& \textbf{PSNR} & \textbf{SSIM} & \textbf{PSNR} & \textbf{SSIM} & \textbf{PSNR} & \textbf{SSIM} & \textbf{PSNR} & \textbf{SSIM} \\
		\midrule
		\multicolumn{9}{@{}l@{}}{\textcolor{gray}{\textit{Set12}}} \\
		ANLM & 14.11 & 27.86 & 17.97 & 40.04 & 23.04 & 60.08 & 22.94 & 66.98 \\
		DIP & 17.08 & 48.69 & 18.18 & 56.81 & 19.04 & 61.86 & 20.83 & 68.97 \\
		DnCNN & 18.36 & 39.91 & 20.72 & 46.02 & 22.43 & 51.99 & 23.44 & 58.01 \\
		SARCAM & 16.25 & 27.15 & 22.56 & 57.74 & 24.29 & 68.34 & 23.41 & 70.61 \\
		SAR2SAR & 17.54 & 45.06 & 21.78 & 55.39 & 22.08 & 59.74 & 22.22 & 61.72 \\
		SARtrans & 18.84 & 51.58 & 19.11 & 52.15 & 18.71 & 52.30 & 18.55 & 52.51 \\
		AGSDNet & 13.46 & 18.71 & 21.29 & 48.10 & \underline{26.14} & \underline{75.33} & \underline{27.76} & \underline{81.23} \\
		SIFSDNet & 18.66 & 38.67 & \underline{23.11} & 61.95 & 25.14 & 72.01 & 25.82 & 76.87 \\
		MONet & \textbf{21.05} & \underline{54.49} & 22.77 & \underline{65.74} & 23.50 & 66.87 & 24.02 & 67.26 \\
		RDDPM & 16.98 & 32.22 & 18.50 & 38.51 & 19.93 & 44.26 & 21.03 & 49.87 \\
		CL-SAR & 15.52 & 29.07 & 16.61 & 55.30 & 17.97 & 71.16 & 19.11 & 77.11 \\
		S3DIP & 16.38 & 50.98 & 17.30 & 56.06 & 17.00 & 62.57 & 16.26 & 64.35 \\
		Ours & \underline{20.58} & \textbf{63.55} & \textbf{23.96} & \textbf{72.14} & \textbf{26.48} & \textbf{77.96} & \textbf{28.43} & \textbf{83.14} \\
		\midrule
		\multicolumn{9}{@{}l@{}}{\textcolor{gray}{\textit{McMaster}}} \\
		ANLM & 11.72 & 22.21 & 13.91 & 30.65 & 16.37 & 39.98 & 19.37 & 50.49 \\
		DIP & 16.09 & 52.75 & 18.70 & 61.90 & 20.29 & 67.68 & 22.46 & 72.67 \\
		DnCNN & 11.83 & 22.86 & 13.86 & 31.98 & 16.56 & 43.12 & 20.08 & 54.77 \\
		SARCAM & 14.33 & 26.44 & 17.78 & 39.64 & 24.08 & 64.60 & 27.43 & 79.20 \\
		SAR2SAR & 13.16 & 35.75 & 20.59 & 54.99 & 22.46 & 64.93 & 23.09 & 68.59 \\
		SARtrans & \underline{18.40} & 52.42 & 19.56 & 53.22 & 19.85 & 53.63 & 19.86 & 53.83 \\
		AGSDNet & 12.18 & 18.11 & 15.14 & 27.49 & 22.66 & 55.87 & \textbf{28.49} & 81.67 \\
		SIFSDNet & 16.01 & 30.62 & 20.04 & 48.33 & \underline{24.90} & 69.54 & \underline{28.33} & \underline{82.02} \\
		MONet & 15.15 & 26.57 & \underline{21.48} & 60.08 & 24.44 & \underline{77.36} & 26.06 & 78.75 \\
		RDDPM & 13.31 & 25.71 & 14.88 & 30.01 & 16.42 & 33.76 & 17.71 & 36.77 \\
		CL-SAR & 15.77 & 36.12 & 17.11 & 60.60 & 18.95 & 76.29 & 20.37 & 81.75 \\
		S3DIP & 14.96 & \underline{55.98} & 15.14 & \underline{64.40} & 15.15 & 64.85 & 15.30 & 62.62 \\
		Ours & \textbf{19.75} & \textbf{63.85} & \textbf{22.98} & \textbf{72.11} & \textbf{25.85} & \textbf{78.79} & 28.20 & \textbf{83.80} \\
		\midrule
		\multicolumn{9}{@{}l@{}}{\textcolor{gray}{\textit{Kodak24}}} \\
		ANLM & 11.29 & 10.03 & 13.49 & 16.32 & 16.13 & 26.33 & 19.27 & 40.46 \\
		DIP & \textbf{19.01} & 50.47 & 20.51 & 55.32 & 22.02 & 59.86 & 23.37 & 64.39 \\
		DnCNN & 11.41 & 10.30 & 13.35 & 16.55 & 16.04 & 27.60 & 19.62 & 43.76 \\
		SARCAM & 13.85 & 15.47 & 16.97 & 26.39 & 22.64 & 51.64 & \underline{25.38} & 69.12 \\
		SAR2SAR & 16.95 & 49.35 & 20.10 & 55.39 & 20.76 & 57.76 & 20.95 & 59.13 \\
		SARtrans & 17.95 & 48.42 & 19.14 & 48.91 & 19.63 & 49.13 & 19.77 & 49.19 \\
		AGSDNet & 11.80 & 10.60 & 14.66 & 17.55 & 21.29 & 43.05 & \textbf{26.30} & 72.83 \\
		SIFSDNet & 15.55 & 19.81 & 19.19 & 36.09 & \underline{23.48} & 58.20 & \textbf{26.30} & 73.14 \\
		MONet & 14.71 & 17.44 & \underline{20.67} & 47.89 & 23.45 & 68.00 & 24.59 & 67.78 \\
		RDDPM & 13.03 & 21.41 & 14.60 & 26.34 & 16.08 & 30.85 & 17.40 & 34.36 \\
		CL-SAR & 15.91 & 23.78 & 18.37 & 50.32 & 20.36 & \textbf{68.74} & 22.36 & \textbf{75.09} \\
		S3DIP & 17.47 & \underline{51.96} & 16.50 & \underline{58.01} & 15.91 & 58.02 & 16.02 & 56.57 \\
		Ours & \underline{18.97} & \textbf{55.55} & \textbf{21.53} & \textbf{61.86} & \textbf{23.65} & \underline{68.16} & 25.35 & \underline{74.01} \\
		\bottomrule[1pt]
	\end{tabularx}
\end{table}

Across the three benchmarks, the method obtains the best value in 18 of the 24 metric and noise-level combinations in Table \ref{tb2}. It leads both metrics from two to eight looks on Set12, both metrics at two looks and PSNR at four looks on Kodak24, and both metrics through four looks on McMaster. At one look, PSNR ranks second on Set12 and Kodak24, whereas SSIM is highest. This difference is consistent with pixelwise fidelity favoring stronger smoothing in the regime where \eqref{eq:cstar} is not identifiable. At eight looks, supervised models obtain higher PSNR on McMaster and Kodak24 by $0.29$ and $0.95$~dB, respectively. The proposed estimator nevertheless gives the best McMaster SSIM. The visual comparisons show that these results arise from retaining edges and texture rather than smoothing alone. Geometry calibration therefore improves consistency across image content and noise levels, although a task-specific learned prior can be more accurate when corruption is weak.

\begin{figure}[t]
	\centering
	\includegraphics[width=\linewidth]{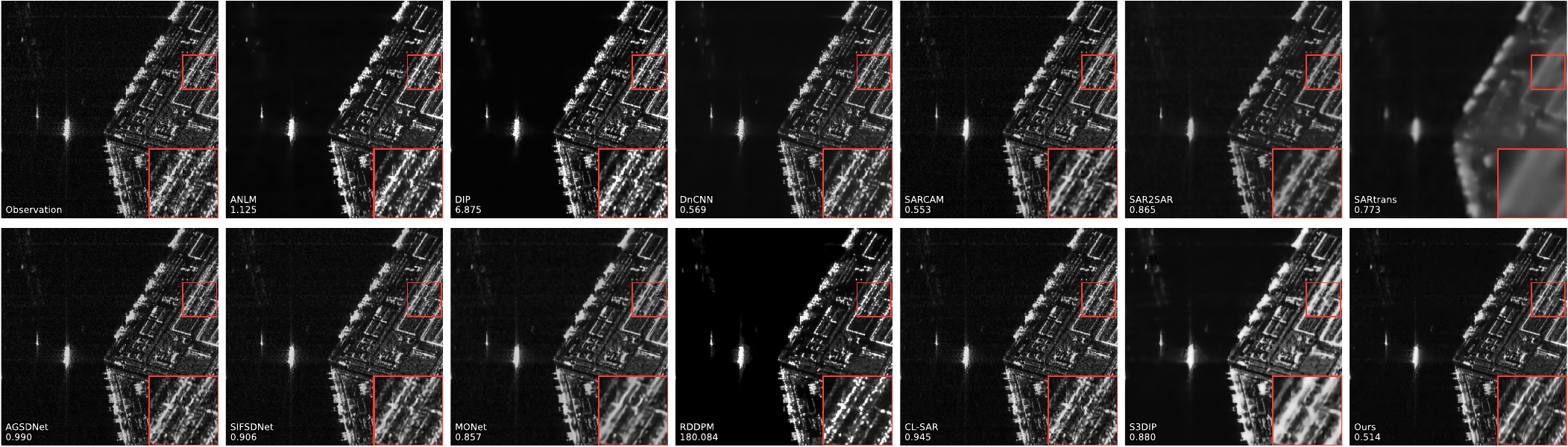}
	\caption{Gaofen-3 results. Panels are ordered as labeled, and insets show the marked regions. Values below the panels are the ratio-image deviations from Table \ref{tb6}.}
	\label{fig_7}
\end{figure}

\begin{figure}[t]
	\centering
	\includegraphics[width=\textwidth]{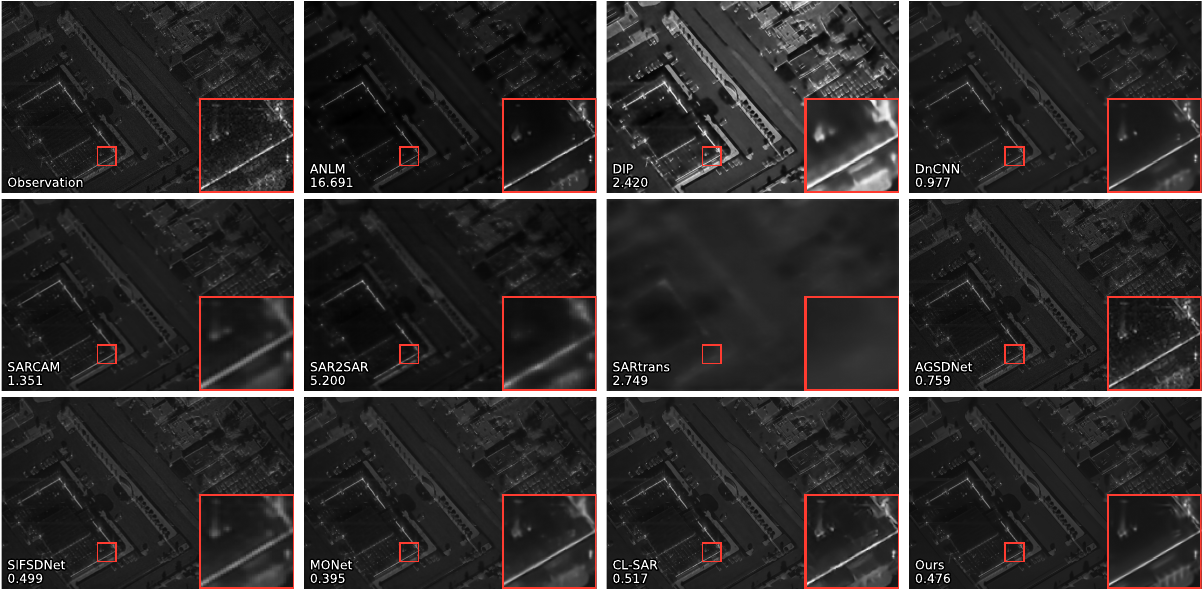}
	\caption{miniSAR results. Insets show the marked regions. Values below the panels are the ratio-image deviations from Table \ref{tb6}.}
	\label{fig_minisar}
\end{figure}

\subsection{Results on Real SAR Imagery}

The real-data study covers urban and natural scenes from six SAR configurations. Representative results for medium-resolution Sentinel-1 and Gaofen-3 and $0.1$~m miniSAR are shown in Figs. \ref{fig_6}, \ref{fig_7}, and \ref{fig_minisar}; Table \ref{tb6} retains the quantitative evaluation for every sensor. The method reduces speckle while preserving linear features and building contours. It uses no sensor-specific training distribution, and sensor dependence enters only through the observation model and estimated look number.

\realresultstable

\begin{figure}[t]
	\centering
	\includegraphics[width=0.9\linewidth]{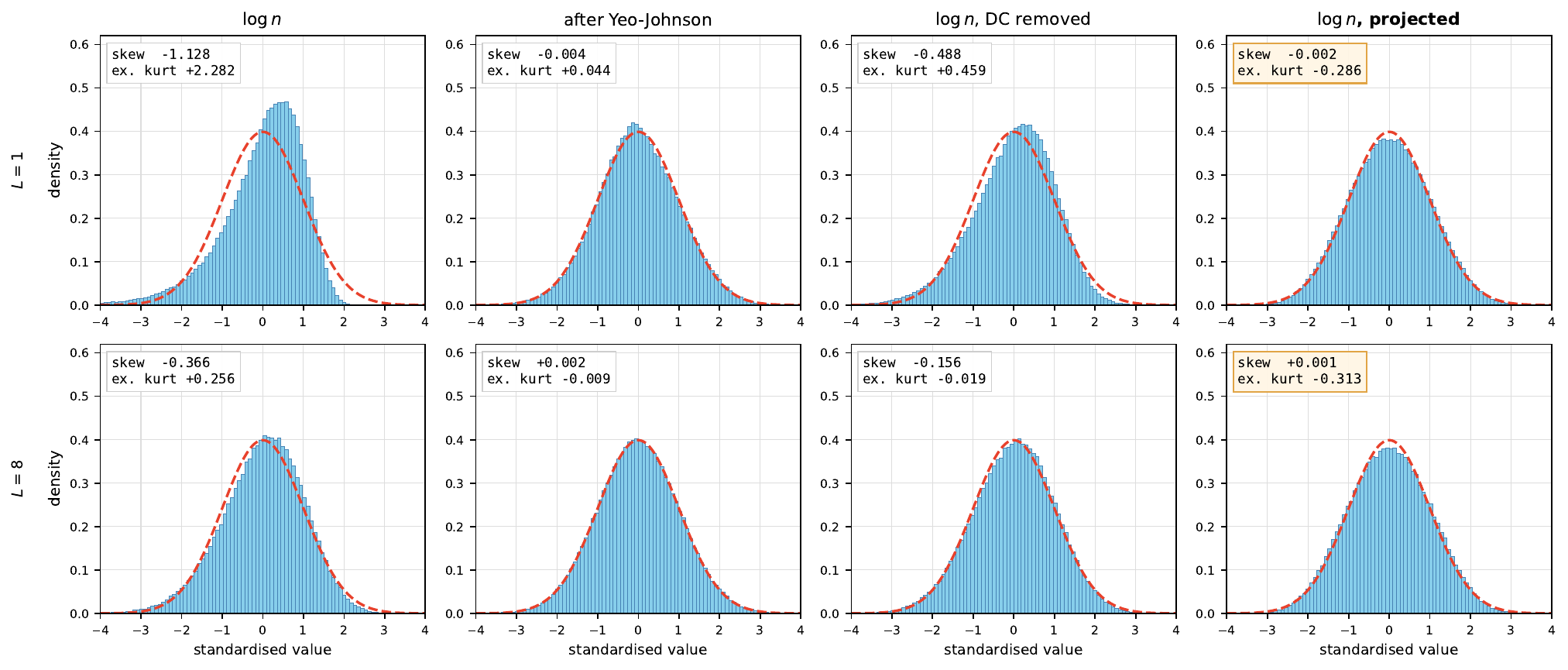}
	\caption{Standardized gamma-noise distributions at one and eight looks in the log domain and after the logarithmic Yeo--Johnson transformation, group-mean removal, and projection. The curve is the standard normal density.}
	\label{fig_13}
\end{figure}

\begin{figure}[t]
	\centering
	\includegraphics[width=0.6\linewidth]{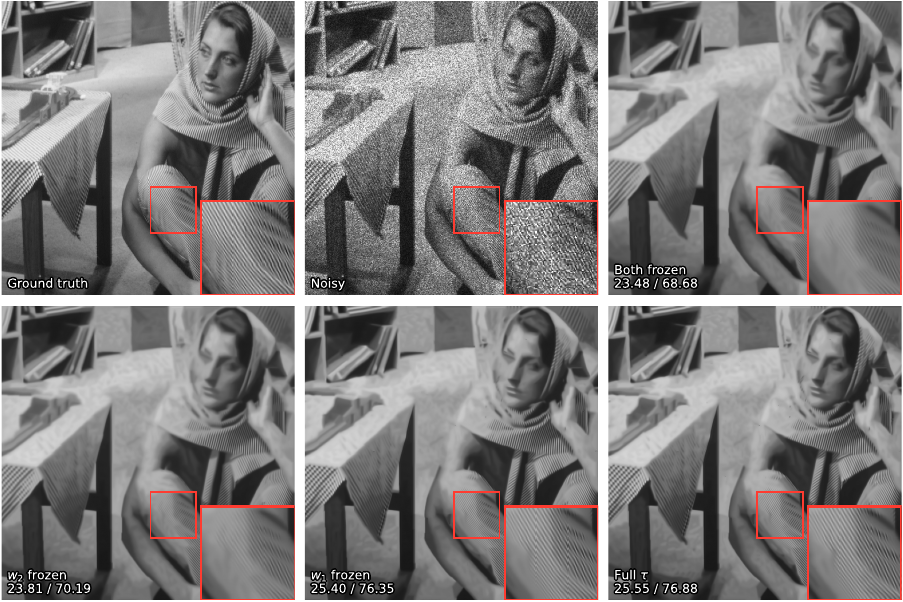}
	\caption{Threshold-factor ablation at four looks. The panels show the reference, observation, both factors fixed, $w_2$ fixed, $w_1$ fixed, and the full model. Insets show texture preservation, and PSNR/SSIM values refer to this scene.}
	\label{fig_14}
\end{figure}

\begin{table}[!t]\setlength{\belowcaptionskip}{6pt}
	\scriptsize
	\centering
	\caption{Module ablation. LYJ denotes the logarithmic Yeo--Johnson transform. The symbols $\checkmark$ and $\times$ indicate enabled and disabled components.\label{tb12}}
	\setlength{\tabcolsep}{0.45pt}
	\begin{tabularx}{\linewidth}{*{8}{>{\centering\arraybackslash}X}}
		\toprule[1pt]
		\multirow{2}{*}{} & \multirow{2}{*}{\textbf{$w_1,w_2$}} & \multirow{2}{*}{\textbf{LYJ}} & \multicolumn{2}{c}{\textbf{Synthetic (4 looks)}} & \multicolumn{3}{c}{\textbf{Real}}  \\
		& & & \multicolumn{1}{c}{\textbf{PSNR}} & \multicolumn{1}{c}{\textbf{SSIM}} & \multicolumn{1}{c}{\textbf{ENL}} & \multicolumn{1}{c}{\textbf{EPI}} & \multicolumn{1}{c}{\textbf{EPD}} \\ 
		\midrule
		(a) & $\times$ & $\times$ & 22.62 & 65.94 & 2.486 & 0.571 & 9.85 \\
		(b) & $\times$ & $\checkmark$ & 23.21 & 67.21 & 2.537 & 0.599 & 10.32 \\
		(c) & $\checkmark$ & $\times$ & \underline{25.60} & \underline{74.79} & \textbf{2.861} & \underline{0.658} & \underline{13.78} \\
		(d) & $\checkmark$ & $\checkmark$ & \textbf{25.98} & \textbf{77.13} & \underline{2.784} & \textbf{0.733} & \textbf{14.33} \\
		\bottomrule[1pt]
	\end{tabularx}
\end{table}

Table \ref{tb6} measures departures of the ratio image from its theoretical mean, variance, and spatial independence. The proposed method performs best on four of the six configurations, ranks second on FARAD-X, and has the lowest mean deviation. It also gives the smallest average deviations in the mean and variance terms. Underfiltering reduces ratio variance, whereas overfiltering transfers image structure and spatial correlation to the ratio; neither can reproduce the expected mean, power, and independence simultaneously. The consistency from medium-resolution Sentinel-1 and Gaofen-3 data to $0.1$~m FARAD imagery supports geometry-calibrated restoration without a sensor-specific prior. Fine miniSAR structures are retained, and its deviation is among the three lowest values.

\begin{figure}[t]
	\centering
	\includegraphics[width=\textwidth]{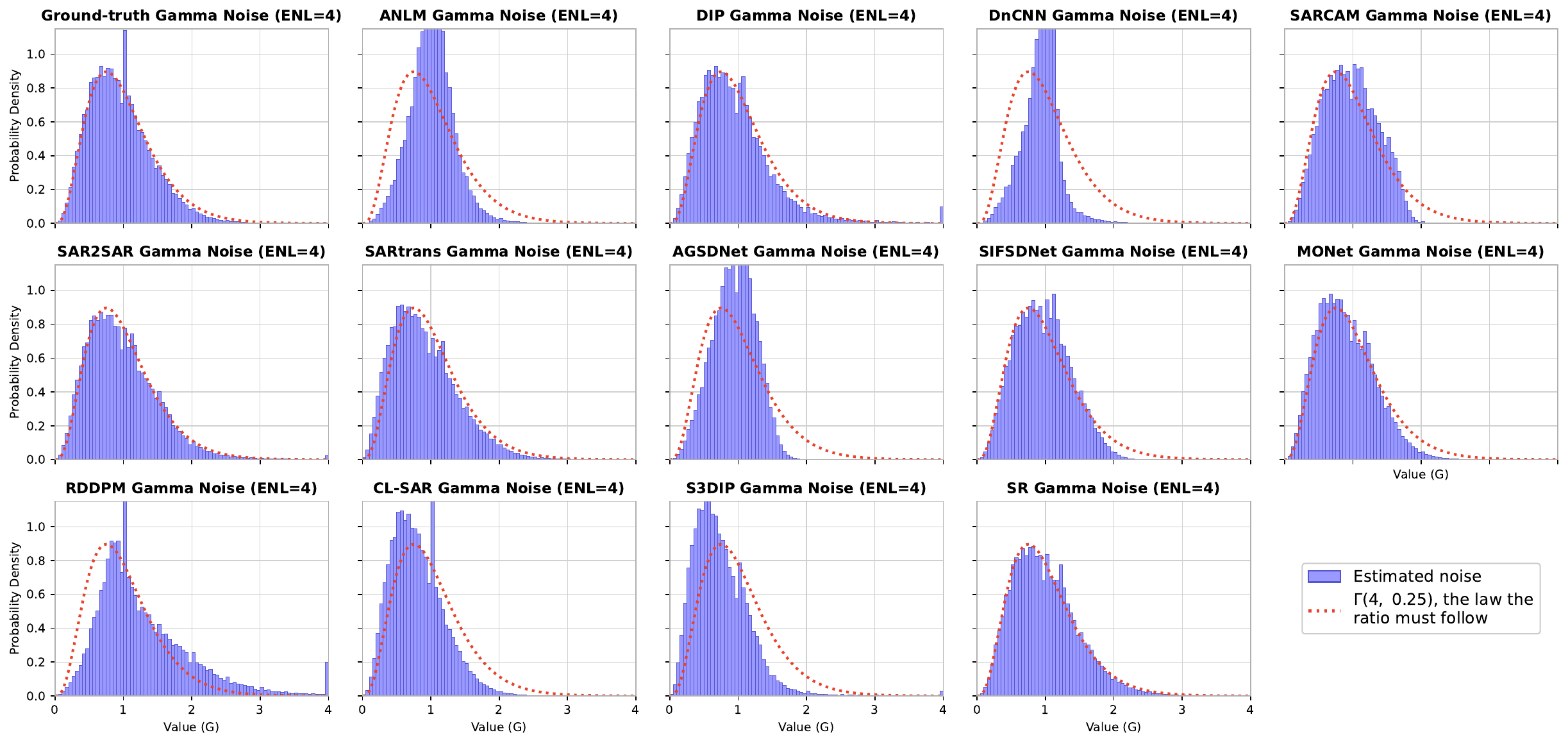}
	\caption{Four-look Kodak24 ratio-image distributions pooled over 24 images. The reference ratio includes quantization effects, and the dotted curve represents ideal gamma noise. Narrower and broader distributions indicate residual noise and structure leakage, respectively.}
	\label{fig_12}
\end{figure}

\begin{table}[!t]\setlength{\belowcaptionskip}{6pt}
	\scriptsize
	\centering
	\caption{Ablation of $\tau_{ik}=w_{2,i}/w_{1,k}^2$. A fixed factor is set to its group mean. Differences between rows isolate the sparse-coding stage.\label{tb13}}
	\setlength{\tabcolsep}{1pt}
	\begin{tabularx}{\linewidth}{p{1.65cm}*{8}{>{\centering\arraybackslash}X}}
		\toprule[1pt]
		\multirow{2}{*}{} & \multicolumn{2}{c}{\textbf{1-Look}} & \multicolumn{2}{c}{\textbf{2-Look}} & \multicolumn{2}{c}{\textbf{4-Look}} & \multicolumn{2}{c}{\textbf{8-Look}}  \\
		& \textbf{PSNR} & \textbf{SSIM} & \textbf{PSNR} & \textbf{SSIM} & \textbf{PSNR} & \textbf{SSIM} & \textbf{PSNR} & \textbf{SSIM} \\
		\midrule
		Full $\tau$ & \textbf{20.30} & 57.87 & \textbf{23.45} & \textbf{67.68} & \textbf{26.01} & \textbf{75.21} & \textbf{28.12} & \textbf{81.11} \\
		$w_1$ frozen & \underline{20.28} & \textbf{58.32} & \underline{23.36} & \underline{67.50} & \underline{25.88} & \underline{74.89} & \underline{27.95} & \underline{80.67} \\
		$w_2$ frozen & 20.10 & 56.92 & 22.66 & 63.91 & 24.62 & 70.39 & 26.49 & 76.61 \\
		Frozen & 20.06 & 56.74 & 22.51 & 63.18 & 24.29 & 69.16 & 26.04 & 75.25 \\
		\bottomrule[1pt]
	\end{tabularx}
\end{table}

\begin{table}[!t]\setlength{\belowcaptionskip}{6pt}
	\scriptsize
	\centering
	\caption{Empirical (M) and predicted (P) $c$ across patch-group geometries.\label{tb14}}
	\renewcommand{\arraystretch}{1.10}
	\setlength{\tabcolsep}{3.3pt}
	\begin{adjustbox}{max width=\linewidth}
	\begin{tabular}{ccc*{6}{c}@{\hspace{13pt}}ccc*{6}{c}}
		\toprule
		\multirow{2}{*}{$p$} & \multirow{2}{*}{$K$} & \multirow{2}{*}{$\gamma$} & \multicolumn{2}{c}{\textbf{2-L}} & \multicolumn{2}{c}{\textbf{4-L}} & \multicolumn{2}{c}{\textbf{8-L}} &
		\multirow{2}{*}{$p$} & \multirow{2}{*}{$K$} & \multirow{2}{*}{$\gamma$} & \multicolumn{2}{c}{\textbf{2-L}} & \multicolumn{2}{c}{\textbf{4-L}} & \multicolumn{2}{c}{\textbf{8-L}} \\
		& & & \textbf{M} & \textbf{P} & \textbf{M} & \textbf{P} & \textbf{M} & \textbf{P} &
		& & & \textbf{M} & \textbf{P} & \textbf{M} & \textbf{P} & \textbf{M} & \textbf{P} \\
		\midrule
		4  & 20 & 0.80 & 1.18 & 1.08 & 0.95 & 0.94 & 0.84 & 0.79 & 8  & 10 & 6.40  & 1.68 & 1.74 & 1.50 & 1.70 & 1.50 & 1.65 \\
		4  & 10 & 1.60 & 1.31 & 1.23 & 1.04 & 1.11 & 0.95 & 0.99 & 12 & 20 & 7.20  & 1.93 & 1.81 & 1.93 & 1.77 & 1.93 & 1.73 \\
		6  & 20 & 1.80 & 1.18 & 1.26 & 1.18 & 1.14 & 1.04 & 1.03 & 10 & 10 & 10.00 & 1.93 & 2.00 & 1.93 & 1.99 & 1.93 & 1.98 \\
		8  & 20 & 3.20 & 1.50 & 1.44 & 1.31 & 1.35 & 1.31 & 1.26 & 16 & 20 & 12.80 & 2.62 & 2.17 & 2.33 & 2.19 & 2.33 & 2.20 \\
		6  & 10 & 3.60 & 1.31 & 1.49 & 1.31 & 1.40 & 1.31 & 1.32 & 12 & 10 & 14.40 & 2.33 & 2.26 & 2.33 & 2.29 & 2.33 & 2.31 \\
		10 & 20 & 5.00 & 1.68 & 1.63 & 1.68 & 1.56 & 1.50 & 1.50 & 16 & 10 & 25.60 & 2.62 & 2.78 & 3.00 & 2.88 & 3.00 & 2.98 \\
		\bottomrule
	\end{tabular}
	\end{adjustbox}
\end{table}

\subsection{Ablation Study}
\label{sec:ablation}

Figure \ref{fig_13} tracks changes in noise skewness and excess kurtosis across the processing stages. At one look, the transformation changes the skewness from $-1.128$ to $-0.004$ and the excess kurtosis from $2.282$ to $0.044$, bringing both statistics close to their Gaussian values of zero. The Kolmogorov--Smirnov distance to normality also decreases from $0.0718$ to $0.0080$. Projection onto the group singular-vector basis also produces an almost symmetric distribution without the transformation: the skewness is $-0.002$, although the excess kurtosis remains negative at $-0.286$. These results show that both transformation and projection reduce noise asymmetry, but do not by themselves establish Gaussianity. Table \ref{tb12} complements this distributional analysis by evaluating how the transformation and adaptive weighting affect restoration performance.

\paragraph{Threshold factors}
Equation \eqref{eq:closed-form} combines the patch-dependent factor $w_1$ and the atom-dependent factor $w_2$ into a single threshold, $\tau_{ik}=w_{2,i}/w_{1,k}^2$. To assess the contribution of each factor's adaptivity, we replace it with its mean within each group while leaving the other factor unchanged. We also test a variant with both factors fixed at their respective group means. This design removes variation across patches or atoms while retaining nonzero shrinkage. Table \ref{tb13} compares these variants within the sparse-coding stage; its absolute scores should not be compared directly with the full-system results in Table \ref{tb2}.

Table \ref{tb13} shows that atom-dependent weighting contributes more to reconstruction accuracy than patch-dependent weighting. Fixing $w_2$ reduces PSNR by $1.39$ and $1.63$~dB at four and eight looks, respectively, whereas fixing $w_1$ reduces it by only $0.13$ and $0.17$~dB. This difference is consistent with their roles. $w_2$ adapts shrinkage to the singular spectrum, while $w_1$ adjusts for reliability differences among matched patches. Patch-dependent weighting remains useful when $w_2$ is fixed, improving PSNR over the both-fixed variant by $0.33$~dB at four looks and $0.45$~dB at eight looks. At one look, however, fixing $w_1$ slightly improves SSIM, indicating that its benefit is not uniform across all metrics and noise levels. Figure \ref{fig_14} illustrates the corresponding visual difference. Periodic texture is better preserved with adaptive $w_2$ than with fixed $w_2$.

\paragraph{Sensitivity}
Table \ref{tb14} compares the predicted correction factor $c$ with its empirical optimum across patch sizes, group sizes, and noise levels. The mean absolute error across the listed configurations is $0.09$. Configurations with similar aspect ratios can have similar optimal corrections. For example, $(p,K)=(8,20)$ and $(6,10)$ give $\gamma=3.2$ and $3.6$, respectively, and both have an empirical optimum of $c=1.31$ at four and eight looks. At eight looks, $(10,20)$ and $(8,10)$ also share the optimum $c=1.50$. These results support using the group aspect ratio to jointly account for patch size and group size when predicting the correction.

We next examine sensitivity to the patch width $p$ and rank-cutoff parameter $r$, updating $c$ according to \eqref{eq:cstar} as the patch-group geometry changes. On McMaster, patch widths of $10$--$12$ pixels give the highest PSNR. Across the tested range of $p=6$--$16$, the PSNR spread is at most $0.48$~dB; at four looks, increasing $p$ from $10$ to $16$ reduces PSNR by $0.37$~dB. Among the tested rank-cutoff settings, $r=1.5$ performs best at two or more looks, while the alternatives reduce PSNR by $0.21$--$0.48$~dB. At one look, $r=1$ improves PSNR by $0.07$~dB relative to $r=1.5$. Based on these results, we use $p=10$ and set $r=1.5$ for two or more looks and $r=1$ for one look.

\subsection{Ratio-Image Statistical Validation}

To assess whether the removed component is consistent with the assumed speckle model, we form the ratio image from the noisy observation $y$ and the restored estimate $\hat{x}$:
\begin{equation}
G_{i,j}=\frac{y_{i,j}}{\hat{x}_{i,j}},
\qquad \hat{x}_{i,j}>0.
\end{equation}
If $\hat{x}$ accurately recovers the underlying image, this ratio approximates the multiplicative noise. We collect the ratios at all valid pixels into a sample set $\{G_n\}$ for distributional analysis.


To determine whether the residual follows the assumed model, we fit a gamma distribution to $\{G_n\}$ by maximum likelihood. Here, $k$ and $\theta$ denote the shape and scale parameters, respectively. We compare the fitted values $\hat{k}$ and $\hat{\theta}$ with the theoretical values $k=L$ and $\theta=1/L$.

\begin{equation}
	(\hat{k}, \hat{\theta}) 
	= \arg\max_{k,\,\theta} \prod_{n} p(G_n; k, \theta)
\end{equation}

Figure \ref{fig_12} and Table \ref{tb5} compare ratio-image statistics on four-look Kodak24. The proposed method gives a ratio mean of $0.9986$, with an absolute deviation from unity of $0.0014$, compared with $0.0178$ for the next-best method, ANLM. Its fitted scale parameter is $0.2370$, also the closest to the theoretical value of $0.25$. The smaller scale parameters of DnCNN and AGSDNet are consistent with narrower ratio distributions and incomplete speckle removal. Conversely, the larger scale and shifted mean of RDDPM are consistent with structural leakage or radiometric bias. Agreement in both mean and scale supports the statistical consistency of the proposed result, although these marginal statistics alone cannot rule out structural leakage. This experiment complements the real-SAR ratio analysis by providing a controlled setting with a known noise level.

\begin{table}[!t]\setlength{\belowcaptionskip}{6pt}
	\scriptsize
	\centering
	\caption{Gamma-fit consistency on four-look Kodak24. The ideal values are $\mathrm{E}[G]=1$ and $\theta=0.25$.}\label{tb5}
	\setlength{\tabcolsep}{4pt}
	\renewcommand{\arraystretch}{0.98}
	\begin{tabular}{lcc@{\hspace{12pt}}lcc}
		\toprule
		\textbf{Method} & \textbf{$\mathrm{E}[G]$} & \textbf{$\theta$} & \textbf{Method} & \textbf{$\mathrm{E}[G]$} & \textbf{$\theta$} \\
		\midrule
		ANLM~\cite{xiao2020asymptotic} & \underline{1.0178} & 0.1085 & DIP~\cite{ulyanov2018deep} & 0.9673 & 0.2973 \\
		DnCNN~\cite{zhang2017beyond} & 0.9591 & 0.0716 & SARCAM~\cite{9633208} & 0.9700 & 0.1790 \\
		SAR2SAR~\cite{dalsasso2021sar2sar} & 0.9817 & 0.2724 & SARtrans~\cite{perera2022transformer} & 0.9050 & \underline{0.2675} \\
		AGSDNet~\cite{thakur2022agsdnet} & 0.9446 & 0.1243 & SIFSDNet~\cite{thakur2022sifsdnet} & 0.9629 & 0.1886 \\
		MONet~\cite{vitale2023sar} & 0.8932 & 0.2047 & RDDPM~\cite{hu2024sar} & 1.3241 & 0.3990 \\
		CL-SAR~\cite{fang2024contrastive} & 0.8025 & 0.1894 & S3DIP~\cite{albisani2025self} & 0.7473 & 0.2162 \\
		\addlinespace[1pt]
		\textbf{Ours} & \textbf{0.9986} & \textbf{0.2370} & & & \\
		\bottomrule
	\end{tabular}
\end{table}


%

\section{Discussion}

\subsection{Implications for Nonlocal Image Modeling}

The two weighting matrices jointly determine a single threshold field, $\tau_{ik}=w_{2,i}/w_{1,k}^{2}$. For fixed groups, orthonormal dictionaries, and weights, coefficient-wise soft thresholding solves the weighted sparse-coding problem exactly, preserving the objective while eliminating the iterative inner solver. The overall restoration algorithm remains iterative because grouping, dictionaries, and noise estimates are updated between outer iterations. Thus, the closed-form result applies to the sparse-coding step rather than the entire restoration procedure.

\subsection{Noise Statistics After Grouped Representation}

The distributional analysis in Section \ref{sec:ablation} supports an approximately Gaussian model for the projected noise coefficients under the tested conditions. However, near-Gaussian marginal statistics do not imply independence from the dictionary, which is estimated from the same noisy group. The effective noise scale therefore still requires calibration, motivating the geometry-based correction used to set the shrinkage thresholds.

\subsection{Reference-free Selection}

Ratio-image statistics are useful for evaluation, but individual statistics can be ambiguous objectives for parameter selection. Matching the marginal distribution does not ensure that image structure is preserved, and weak spatial correlation alone does not establish adequate speckle removal. In our synthetic experiments, correlation-based selection overestimates $c$ and reduces PSNR by $0.39$--$0.53$~dB at two to eight looks. The FARAD results likewise show that weak structure in the ratio image can accompany insufficient filtering. These observations motivate setting $b$ to a fixed value and determining $c$ from the group geometry, rather than optimizing either parameter against a single ratio statistic. We therefore use ratio mean, variance, and spatial correlation as complementary measures of restoration quality.

\section{Conclusion}

We presented a training-free nonlocal estimator for image denoising under gamma-distributed multiplicative noise. An orthonormal singular-vector dictionary yields an exact soft-thresholding solution to the weighted Lasso, eliminating the iterative inner solver. A random-matrix correction links the shrinkage scale to the patch-group aspect ratio $\gamma=p^2/K$. The distributional analysis supports an approximately Gaussian model for projected noise coefficients under the tested conditions, while the geometry-based correction addresses the noise dependence of the adaptive dictionary. On Set12, McMaster, and Kodak24, the method achieves the best result in 18 of 24 PSNR/SSIM comparisons with twelve published methods. It also achieves the lowest mean ratio-image deviation across six real SAR configurations from five sensors. Together, these results support the effectiveness of geometry-calibrated nonlocal shrinkage for reconstruction fidelity and structure preservation. Future work will examine correlated noise and extend the framework to complex-valued and multitemporal SAR observations.

\bibliographystyle{elsarticle-num}
\bibliography{reference}

@inproceedings{hu2024sar,
	title={SAR despeckling via regional denoising diffusion probabilistic model},
	author={Hu, Xuran and Xu, Ziqiang and Chen, Zhihan and Feng, Zhenpeng and Zhu, Mingzhe and Stankovi{\'c}, Ljubi{\v{s}}a},
	booktitle={2024 IEEE International Geoscience and Remote Sensing Symposium (IGARSS)},
	pages={7226--7230},
	year={2024},
	organization={IEEE}
}

@article{xu2023edge,
	title={Edge preserved low-rank SAR image despeckling via hierarchical prior knowledge regulation},
	author={Xu, Zhiyong and Feng, Xiaolin and Tian, Sirui and Shen, Xiang-Jun and Zhang, Hong and Wang, Chao},
	journal={IEEE Transactions on Geoscience and Remote Sensing},
	volume={61},
	pages={1--17},
	year={2023},
	publisher={IEEE}
}

@article{ma2024despeckling,
	title={Despeckling SAR Images with Log-Yeo-Johnson Transformation and Conditional Diffusion Models},
	author={Ma, Yaobin and Ke, Peng and Aghababaei, Hossein and Chang, Ling and Wei, Jingbo},
	journal={IEEE Transactions on Geoscience and Remote Sensing},
	volume={62},
	pages={5215417},
	year={2024},
	doi={10.1109/TGRS.2024.3419083},
	publisher={IEEE}
}

@article{aharon2006k,
	title={K-SVD: An algorithm for designing overcomplete dictionaries for sparse representation},
	author={Aharon, Michal and Elad, Michael and Bruckstein, Alfred},
	journal={IEEE Transactions on signal processing},
	volume={54},
	number={11},
	pages={4311--4322},
	year={2006},
	publisher={IEEE}
}

@article{xiao2020asymptotic,
	title={Asymptotic non-local means image denoising algorithm},
	author={Xing, Xiao-Xiao and Wang, Hai-Long and Li, Jian and Zhang, Xuan-De},
	journal={Acta Automatica Sinica},
	volume={46},
	number={9},
	pages={1952--1960},
	year={2020}
}

@article{zhang2017beyond,
	title={Beyond a gaussian denoiser: Residual learning of deep cnn for image denoising},
	author={Zhang, Kai and Zuo, Wangmeng and Chen, Yunjin and Meng, Deyu and Zhang, Lei},
	journal={IEEE transactions on image processing},
	volume={26},
	number={7},
	pages={3142--3155},
	year={2017},
	publisher={IEEE}
}

@ARTICLE{9633208,
	author={Ko, Jaekyun and Lee, Sanghwan},
	journal={IEEE Journal of Selected Topics in Applied Earth Observations and Remote Sensing}, 
	title={SAR Image Despeckling Using Continuous Attention Module}, 
	year={2022},
	volume={15},
	number={},
	pages={3-19},
	doi={10.1109/JSTARS.2021.3132027}}

@article{dalsasso2021sar2sar,
	title={SAR2SAR: A semi-supervised despeckling algorithm for SAR images},
	author={Dalsasso, Emanuele and Denis, Lo{\"\i}c and Tupin, Florence},
	journal={IEEE Journal of Selected Topics in Applied Earth Observations and Remote Sensing},
	volume={14},
	pages={4321--4329},
	year={2021},
	publisher={IEEE}
}

@inproceedings{thakur2022sifsdnet,
	title={SIFSDNet: Sharp image feature based SAR denoising network},
	author={Thakur, Ramesh Kumar and Maji, Suman Kumar},
	booktitle={IGARSS 2022-2022 IEEE International Geoscience and Remote Sensing Symposium},
	pages={3428--3431},
	year={2022},
	organization={IEEE}
}

@article{thakur2022agsdnet, title={AGSDNet: Attention and Gradient-Based SAR Denoising Network}, author={Thakur, Ramesh Kumar and Maji, Suman Kumar}, journal={IEEE Geoscience and Remote Sensing Letters}, volume={19}, pages={1--5}, year={2022}, publisher={IEEE} }

@article{vitale2023sar,
	title={SAR despeckling using multiobjective neural network trained with generic statistical samples},
	author={Vitale, Sergio and Ferraioli, Giampaolo and Frery, Alejandro C and Pascazio, Vito and Yue, Dong-Xiao and Xu, Feng},
	journal={IEEE Transactions on Geoscience and Remote Sensing},
	volume={61},
	pages={1--12},
	year={2023},
	publisher={IEEE}
}

@ARTICLE{10032489,
	author={Baraha, Satyakam and Sahoo, Ajit Kumar},
	journal={IEEE Geoscience and Remote Sensing Letters}, 
	title={Speckle Removal Using Dictionary Learning and PnP-Based Fast Iterative Shrinkage Threshold Algorithm}, 
	year={2023},
	volume={20},
	number={},
	pages={1-5},
	doi={10.1109/LGRS.2023.3241191}}

@article{liu2017over,
	title={An over-complete dictionary design based on GSR for SAR image despeckling},
	author={Liu, Su and Zhang, Gong and Soon, Yeo Tat},
	journal={IEEE Geoscience and Remote Sensing Letters},
	volume={14},
	number={12},
	pages={2230--2234},
	year={2017},
	publisher={IEEE}
}

@article{9484779,
	author={Zhang, Junchao and Chen, Jianlai and Yu, Hanwen and Yang, Degui and Xu, Xiaoqing and Xing, Mengdao},
	journal={IEEE Journal of Selected Topics in Applied Earth Observations and Remote Sensing}, 
	title={Learning an SAR Image Despeckling Model Via Weighted Sparse Representation}, 
	year={2021},
	volume={14},
	number={},
	pages={7148-7158},
	doi={10.1109/JSTARS.2021.3097119}}

@article{gavish2014optimal,
	author={Gavish, Matan and Donoho, David L.},
	journal={IEEE Transactions on Information Theory},
	title={The Optimal Hard Threshold for Singular Values is $4/\sqrt{3}$},
	year={2014},
	volume={60},
	number={8},
	pages={5040--5053},
	doi={10.1109/TIT.2014.2323359}}

@inproceedings{perera2022transformer,
	title={Transformer-based SAR image despeckling},
	author={Perera, Malsha V and Bandara, Wele Gedara Chaminda and Valanarasu, Jeya Maria Jose and Patel, Vishal M},
	booktitle={2022 IEEE International Geoscience and Remote Sensing Symposium (IGARSS)},
	pages={751--754},
	year={2022},
	organization={IEEE}
}

@article{pan2024sar,
	title={SAR image despeckling based on denoising diffusion probabilistic model and swin transformer},
	author={Pan, Yucheng and Zhong, Liheng and Chen, Jingdong and Li, Heping and Zhang, Xianlong and Pan, Bin},
	journal={Remote Sensing},
	volume={16},
	number={17},
    pages={3222},
    doi={10.3390/rs16173222},
	year={2024}
}

@article{molini2021speckle2void,
	title={Speckle2Void: Deep self-supervised SAR despeckling with blind-spot convolutional neural networks},
	author={Bordone Molini, Andrea and Valsesia, Diego and Fracastoro, Giulia and Magli, Enrico},
	journal={IEEE Transactions on Geoscience and Remote Sensing},
	volume={60},
	pages={1--17},
	year={2022},
	publisher={IEEE}
}

@article{kato2024polmerlin,
	title={PolMERLIN: Self-supervised polarimetric complex SAR image despeckling with masked networks},
	author={Kato, Shunya and Saito, Masaki and Ishiguro, Katsuhiko and Cummings, Sol},
	journal={IEEE Geoscience and Remote Sensing Letters},
	volume={21},
	pages={1--5},
	year={2024},
	publisher={IEEE}
}

@article{torres2012gmes,
	title={GMES Sentinel-1 mission},
	author={Torres, Ramon and Snoeij, Paul and Geudtner, Dirk and Bibby, David and Davidson, Malcolm and Attema, Evert and Potin, Pierre and Rommen, Bj{\"O}rn and Floury, Nicolas and Brown, Mike and others},
	journal={Remote sensing of environment},
	volume={120},
	pages={9--24},
	year={2012},
	publisher={Elsevier}
}

@article{zhao2021china,
	title={China's Gaofen-3 satellite system and its application and prospect},
	author={Zhao, Liangbo and Zhang, Qingjun and Li, Yan and Qi, Yalin and Yuan, Xinzhe and Liu, Jie and Li, Hangliang},
	journal={IEEE Journal of Selected Topics in Applied Earth Observations and Remote Sensing},
	volume={14},
	pages={11019--11028},
	year={2021},
	publisher={IEEE}
}

@article{fang2024contrastive,
	title={Contrastive learning for real SAR image despeckling},
	author={Fang, Yangtian and Liu, Rui and Peng, Yini and Guan, Jianjun and Li, Duidui and Tian, Xin},
	journal={ISPRS Journal of Photogrammetry and Remote Sensing},
	volume={218},
	pages={376--391},
	year={2024},
	publisher={Elsevier}
}

@article{albisani2025self,
	title={Self-supervised {SAR} despeckling using deep image prior},
	author={Albisani, Chiara and Baracchi, Daniele and Piva, Alessandro and Argenti, Fabrizio},
	journal={Pattern Recognition Letters},
	volume={190},
	pages={169--176},
	year={2025},
	doi={10.1016/j.patrec.2025.02.021}
}

@inproceedings{ulyanov2018deep,
	title={Deep image prior},
	author={Ulyanov, Dmitry and Vedaldi, Andrea and Lempitsky, Victor},
	booktitle={Proceedings of the IEEE conference on computer vision and pattern recognition},
	pages={9446--9454},
	year={2018}
}

@article{lin2025speckle2self,
	title={Speckle2Self: Learning Self-Supervised Despeckling with Attention Mechanism for SAR Images},
	author={Lin, Huiping and Su, Xin and Zeng, Zhiqiang and Xing, Cheng and Yin, Junjun},
	journal={Remote Sensing},
	volume={17},
	number={23},
	pages={3840},
	year={2025},
	publisher={MDPI}
}

@article{denis2025just,
	title={Just project! Multichannel despeckling, the easy way},
	author={Denis, Lo{\"\i}c and Dalsasso, Emanuele and Tupin, Florence},
	journal={IEEE Transactions on Geoscience and Remote Sensing},
	volume={63},
	pages={1--11},
	year={2025},
	publisher={IEEE}
}

@article{guo2025efficient,
	title={Efficient Conditional Diffusion Model for SAR Despeckling},
	author={Guo, Zhenyu and Hu, Weidong and Zheng, Shichao and Zhang, Binchao and Zhou, Ming and Peng, Jincheng and Yao, Zhiyu and Feng, Minghao},
	journal={Remote Sensing},
	volume={17},
	number={17},
	pages={2970},
	year={2025},
	publisher={MDPI}
}

@article{ran2025tunable,
	title={A tunable despeckling neural network stabilized via diffusion equation},
	author={Ran, Yi and Guo, Zhichang and Li, Jia and Li, Yao and Burger, Martin and Wu, Boying},
	journal={Signal Processing},
	volume={239},
	pages={110324},
	year={2026},
	doi={10.1016/j.sigpro.2025.110324},
	publisher={Elsevier}
}

@article{yan2025nonlocal,
	title={Nonlocal Matrix Rank Minimization Method for Multiplicative Noise Removal},
	author={Yan, Hui-Yin},
	journal={Communications on Applied Mathematics and Computation},
	volume={7},
	number={5},
	pages={1744--1768},
	year={2025},
	doi={10.1007/s42967-024-00396-9}
}

@article{saha2025cdcfrn,
	title={{SAR-CDCFRN}: A Novel {SAR} Despeckling Approach Utilizing Correlated Dual Channel Feature-Based Residual Network},
	author={Saha, Anirban and Arihant, K. R. and Maji, Suman Kumar},
	journal={Signal Processing: Image Communication},
	volume={133},
	pages={117267},
	year={2025},
	doi={10.1016/j.image.2025.117267}
}

@article{wang2025dps,
	title={Diffusion Posterior Sampling for {SAR} Despeckling},
	author={Wang, Zelong and Han, Jialing and Zhang, Chenlin},
	journal={IEEE Transactions on Geoscience and Remote Sensing},
	volume={63},
	pages={5204619},
	year={2025},
	doi={10.1109/TGRS.2025.3541013}
}

@article{lu2025dispeckle,
	title={{DiSpeckle}: Diffusion Model That Unwinds Speckle Formation With Off-the-Shelf Gaussian Denoisers},
	author={Lu, Danwei and Liu, Chao and Yin, Junjun and Yang, Jian},
	journal={IEEE Transactions on Geoscience and Remote Sensing},
	volume={63},
	pages={5224622},
	year={2025},
	doi={10.1109/TGRS.2025.3630133}
}

@article{chen2026sds,
	title={Self-Supervised Despeckling Based Solely on {SAR} Intensity Images: A General Strategy},
	author={Chen, Liang and Yin, Yifei and Shi, Hao and He, Jingfei and Li, Wei},
	journal={ISPRS Journal of Photogrammetry and Remote Sensing},
	volume={231},
	pages={854--873},
	year={2026},
	doi={10.1016/j.isprsjprs.2025.11.025}
}

@article{yang2026glc,
	title={Self-Supervised Global-Local Collaborative Network for Real {SAR} Despeckling},
	author={Yang, Yang and Xu, Jiangong and Bai, Yuchuan and Chen, Liangyu and Li, Junli and Pan, Jun and Wang, Mi},
	journal={International Journal of Applied Earth Observation and Geoinformation},
	volume={146},
	pages={105135},
	year={2026},
	doi={10.1016/j.jag.2026.105135}
}

@article{dabov2007bm3d,
  author={Dabov, Kostadin and Foi, Alessandro and Katkovnik, Vladimir and Egiazarian, Karen},
  journal={IEEE Transactions on Image Processing},
  title={Image Denoising by Sparse 3-D Transform-Domain Collaborative Filtering},
  year={2007},
  volume={16},
  number={8},
  pages={2080--2095},
  doi={10.1109/TIP.2007.901238}
}

@inproceedings{mairal2009nonlocal,
  author={Mairal, Julien and Bach, Francis and Ponce, Jean and Sapiro, Guillermo and Zisserman, Andrew},
  title={Non-Local Sparse Models for Image Restoration},
  booktitle={2009 IEEE 12th International Conference on Computer Vision},
  year={2009},
  pages={2272--2279},
  doi={10.1109/ICCV.2009.5459452}
}

@inproceedings{gu2014wnnm,
  author={Gu, Shuhang and Zhang, Lei and Zuo, Wangmeng and Feng, Xiangchu},
  title={Weighted Nuclear Norm Minimization with Application to Image Denoising},
  booktitle={Proceedings of the IEEE Conference on Computer Vision and Pattern Recognition},
  year={2014},
  pages={2862--2869}
}

@inproceedings{lehtinen2018noise2noise,
  author={Lehtinen, Jaakko and Munkberg, Jacob and Hasselgren, Jon and Laine, Samuli and Karras, Tero and Aittala, Miika and Aila, Timo},
  title={Noise2Noise: Learning Image Restoration without Clean Data},
  booktitle={Proceedings of the 35th International Conference on Machine Learning},
  series={Proceedings of Machine Learning Research},
  volume={80},
  pages={2965--2974},
  year={2018},
  publisher={PMLR}
}

@inproceedings{krull2019noise2void,
  author={Krull, Alexander and Buchholz, Tim-Oliver and Jug, Florian},
  title={Noise2Void---Learning Denoising From Single Noisy Images},
  booktitle={Proceedings of the IEEE/CVF Conference on Computer Vision and Pattern Recognition},
  year={2019},
  pages={2129--2137}
}

@inproceedings{quan2020self2self,
  author={Quan, Yuhui and Chen, Mingqin and Pang, Tongyao and Ji, Hui},
  title={Self2Self With Dropout: Learning Self-Supervised Denoising From Single Image},
  booktitle={Proceedings of the IEEE/CVF Conference on Computer Vision and Pattern Recognition},
  year={2020},
  pages={1890--1898}
}

@inproceedings{wang2022blind2unblind,
  author={Wang, Zejin and Liu, Jiazheng and Li, Guoqing and Han, Hua},
  title={Blind2Unblind: Self-Supervised Image Denoising With Visible Blind Spots},
  booktitle={Proceedings of the IEEE/CVF Conference on Computer Vision and Pattern Recognition},
  year={2022},
  pages={2027--2036}
}

@inproceedings{li2025positive2negative,
  author={Li, Tong and Wang, Lizhi and Xu, Zhiyuan and Zhu, Lin and Lu, Wanxuan and Huang, Hua},
  title={Positive2Negative: Breaking the Information-Lossy Barrier in Self-Supervised Single Image Denoising},
  booktitle={Proceedings of the IEEE/CVF Conference on Computer Vision and Pattern Recognition},
  year={2025},
  pages={17924--17934}
}

@article{deledalle2017mulog,
  author={Deledalle, Charles-Alban and Denis, Lo{\"i}c and Tabti, Sonia and Tupin, Florence},
  journal={IEEE Transactions on Image Processing},
  title={MuLoG, or How to Apply Gaussian Denoisers to Multi-Channel SAR Speckle Reduction?},
  year={2017},
  volume={26},
  number={9},
  pages={4389--4403},
  doi={10.1109/TIP.2017.2713946}
}

@article{parrilli2012sar,
  author={Parrilli, Sara and Poderico, Mariana and Angelino, Cesario Vincenzo and Verdoliva, Luisa},
  journal={IEEE Transactions on Geoscience and Remote Sensing},
  title={A Nonlocal SAR Image Denoising Algorithm Based on LLMMSE Wavelet Shrinkage},
  year={2012},
  volume={50},
  number={2},
  pages={606--616},
  doi={10.1109/TGRS.2011.2161586}
}

@article{baraha2022systematic,
  author={Baraha, Satyakam and Sahoo, Ajit Kumar and Modalavalasa, Sowjanya},
  title={A systematic review on recent developments in nonlocal and variational methods for {SAR} image despeckling},
  journal={Signal Processing},
  volume={196},
  pages={108521},
  year={2022},
  doi={10.1016/j.sigpro.2022.108521},
  publisher={Elsevier}
}

@article{li2020statistical,
  author={Li, Yu and Wang, Shuyun and Zhao, Quanhua and Wang, Guanghui},
  title={A new {SAR} image filter for preserving speckle statistical distribution},
  journal={Signal Processing},
  volume={176},
  pages={107706},
  year={2020},
  doi={10.1016/j.sigpro.2020.107706},
  publisher={Elsevier}
}

@article{xu2017fncsr,
  author={Xu, Shaoping and Yang, Xiaohui and Jiang, Shunliang},
  title={A fast nonlocally centralized sparse representation algorithm for image denoising},
  journal={Signal Processing},
  volume={131},
  pages={99--112},
  year={2017},
  doi={10.1016/j.sigpro.2016.08.006},
  publisher={Elsevier}
}

@article{yuan2024lrenss,
  author={Yuan, Wei and Liu, Han and Liang, Lili and Wang, Wenqing and Liu, Ding},
  title={Image restoration via joint low-rank and external nonlocal self-similarity prior},
  journal={Signal Processing},
  volume={215},
  pages={109284},
  year={2024},
  doi={10.1016/j.sigpro.2023.109284},
  publisher={Elsevier}
}

@article{gavaskar2023pnp,
  author={Gavaskar, Ruturaj G. and Athalye, Chirayu D. and Chaudhury, Kunal N.},
  title={On exact and robust recovery for plug-and-play compressed sensing},
  journal={Signal Processing},
  volume={211},
  pages={109100},
  year={2023},
  doi={10.1016/j.sigpro.2023.109100},
  publisher={Elsevier}
}

@article{kang2023logsar,
  author={Kang, Jian and Ji, Tengyu and Zhang, Zhe and Fernandez-Beltran, Ruben},
  title={{SAR} time series despeckling via nonlocal matrix decomposition in logarithm domain},
  journal={Signal Processing},
  volume={209},
  pages={109040},
  year={2023},
  doi={10.1016/j.sigpro.2023.109040},
  publisher={Elsevier}
}

\end{document}